\documentclass{article}

\PassOptionsToPackage{numbers, compress}{natbib}

\usepackage[preprint]{neurips_2024}

\usepackage{geometry}

\usepackage{tikz}
\usepackage{xcolor}
\usepackage[utf8]{inputenc}
\usepackage[T1]{fontenc}
\usepackage{url}
\usepackage{microtype}
\usepackage{graphicx}
\usepackage{booktabs}
\usepackage{amsfonts}
\usepackage{amsmath}
\usepackage{amssymb}
\usepackage{nicefrac}
\usepackage{xcolor}
\usepackage[table]{xcolor}
\usepackage{array}
\usepackage{tabularx}
\usepackage{subcaption}
\usepackage{wrapfig}
\usepackage{enumitem}
\usepackage{multirow}
\usepackage{makecell}
\usepackage{longtable}
\usepackage[figuresleft]{rotating}
\usepackage{listings}
\usepackage[export]{adjustbox}
\usepackage[tableposition=top]{caption}
\usepackage{colortbl}
\usepackage{color}
\usepackage{pifont}
\usepackage{fontawesome5}
\usepackage{hyperref}
\usepackage{ragged2e}
\usepackage{titlesec}
\usepackage{titletoc}
\usepackage{fancyhdr}
\usepackage[most]{tcolorbox}
\usepackage{tikz}
\usetikzlibrary{arrows.meta,positioning,calc,shapes,fit,decorations.pathreplacing}

\definecolor{themeblue}{RGB}{44,92,168}
\definecolor{themedark}{RGB}{32,52,84}
\definecolor{themelight}{RGB}{238,244,252}
\definecolor{headergray}{RGB}{243,245,247}
\definecolor{rowlight}{RGB}{250,250,252}
\definecolor{tocgray}{RGB}{95,95,95}
\definecolor{green}{RGB}{80,204,72}
\definecolor{accentred}{RGB}{196,72,72}
\usepackage{tikz}
\usetikzlibrary{positioning,arrows.meta,calc}
\usepackage[most]{tcolorbox}
\usepackage{adjustbox}
\usepackage{xcolor}

\definecolor{cardgold}{HTML}{B08B57}
\definecolor{cardblue}{HTML}{4C78A8}
\definecolor{cardteal}{HTML}{4C9A8A}
\definecolor{cardpurple}{HTML}{7A6FA6}

\hypersetup{
    colorlinks=true,
    linkcolor=themeblue,
    citecolor=themeblue,
    urlcolor=themeblue
}

\renewcommand{\arraystretch}{1.12}
\newcolumntype{C}[1]{>{\centering\arraybackslash}p{#1}}
\newcolumntype{L}[1]{>{\arraybackslash}p{#1}}

\titleformat{\section}
  {\Large\bfseries\color{themedark}}
  {\makebox[1.7em][c]{\color{themeblue}\thesection}}
  {0.75em}
  {}
  [\vspace{0.3ex}\color{themeblue!45}\titlerule]

\titleformat{\subsection}
  {\large\bfseries\color{themeblue}}
  {\thesubsection}{0.65em}
  {}

\titleformat{\subsubsection}
  {\normalsize\bfseries\color{themedark}}
  {\thesubsubsection}{0.6em}
  {}

\titlespacing*{\section}{0pt}{1.5ex plus .2ex}{1.0ex}
\titlespacing*{\subsection}{0pt}{1.0ex plus .2ex}{0.6ex}
\titlespacing*{\subsubsection}{0pt}{0.8ex plus .2ex}{0.4ex}

\titlecontents{section}
  [0em]
  {\vspace{0.25em}\bfseries\color{themeblue}}
  {\thecontentslabel\hspace{0.6em}}
  {}
  {\hfill\color{themeblue}\contentspage}

\titlecontents{subsection}
  [1.8em]
  {\small\color{tocgray}}
  {\thecontentslabel\hspace{0.6em}}
  {}
  {\hfill\color{tocgray}\contentspage}

\titlecontents{subsubsection}
  [3.8em]
  {\small\itshape\color{tocgray}}
  {\thecontentslabel\hspace{0.6em}}
  {}
  {\hfill\color{tocgray}\contentspage}

\fancyheadoffset[L]{0pt}
\fancyheadoffset[R]{0pt}

\usepackage{tikz}
\usetikzlibrary{calc, positioning, shadows}
\usepackage{adjustbox}
\usepackage{xcolor}
\usepackage[scaled]{helvet}
\definecolor{rhline}{RGB}{46,84,130}
\definecolor{rhcat}{RGB}{241,236,223}
\definecolor{rhref}{RGB}{223,235,245}
\definecolor{rhtext}{RGB}{30,30,30}
\definecolor{rhcite}{RGB}{18,56,140}

\title{
Reward Hacking in the Era of Large Models: Mechanisms, Emergent Misalignment, Challenges
}

\author{%
  Xiaohua Wang\thanks{Core contributors.},
  Muzhao Tian$^*$,
  Yuqi Zeng$^*$,
  Zisu Huang$^*$,
  Jiakang Yuan$^*$, \\
  \textbf{Bowen Chen$^*$,}
  \textbf{Jingwen Xu$^*$,}
  \textbf{Mingbo Zhou$^*$,} \\
  \textbf{Wenhao Liu,}
  \textbf{Muling Wu, }
  \textbf{Zhengkang Guo, }
  \textbf{Qi Qian,}
  \textbf{Feiran Zhang, }
  \textbf{Ruicheng Yin,} \\
  \textbf{Changze Lv, }
  \textbf{Tao chen, }
  \textbf{Xiaoqing Zheng}\thanks{Correspondence to: \texttt{zhengxq@fudan.edu.cn}},
  \textbf{Xuanjing Huang} \\
  \\
  \textbf{Fudan NLP Group} \\
}

\begin{document}

\vspace*{-2em} 

\begin{tcolorbox}[
    colback=gray!8,         
    colframe=white,         
    width=\textwidth,       
    arc=3mm,                
    boxrule=0pt,            
    left=15pt, right=15pt, top=18pt, bottom=15pt,  
    drop shadow=black!10,    
    enhanced                 
]
 
\fontfamily{qpl}\selectfont

{\LARGE \bfseries \color{themedark} Reward Hacking in the Era of Large Models: \par}
\vspace{6pt}
{\Large \itshape \color{themedark} Mechanisms, Emergent Misalignment, Challenges \par}
\vspace{1.5em}

{\normalsize \bfseries
  Xiaohua Wang$^*$, Muzhao Tian$^*$, Yuqi Zeng$^*$, Zisu Huang$^*$, Jiakang Yuan$^*$, 
  Bowen Chen$^*$, Jingwen Xu$^*$, Mingbo Zhou$^*$, Wenhao Liu, Muling Wu, Zhengkang Guo, 
  Qi Qian, Yifei Wang, Feiran Zhang, Ruicheng Yin, Shihan Dou, Changze Lv, Tao Chen, Kaitao Song, Xu Tan, Tao Gui, Xiaoqing Zheng$^{\ddagger}$, Xuanjing Huang \par
}
\vspace{1em}

{\small
  $^*$Core contributors. \quad 
  $^\ddagger$Correspondence. \par
  \vspace{0.4em}
  \textbf{Affiliations:} Fudan NLP Group \par
}
\vspace{1.5em}

{\small \color{black!90}
Reinforcement Learning from Human Feedback (RLHF) and related alignment paradigms have become central to steering large language models (LLMs) and multimodal large language models (MLLMs) toward human-preferred behaviors. However, these approaches introduce a systemic vulnerability: reward hacking, where models exploit imperfections in learned reward signals to maximize proxy objectives without fulfilling true task intent. As models scale and optimization intensifies, such exploitation manifests as verbosity bias, sycophancy, hallucinated justification, benchmark overfitting, and, in multimodal settings, perception--reasoning decoupling and evaluator manipulation. Recent evidence further suggests that seemingly benign shortcut behaviors can generalize into broader forms of misalignment, including deception and strategic gaming of oversight mechanisms. In this survey, we propose the Proxy Compression Hypothesis (PCH) as a unifying framework for understanding reward hacking. We formalize reward hacking as an emergent consequence of optimizing expressive policies against compressed reward representations of high-dimensional human objectives. Under this view, reward hacking arises from the interaction of objective compression, optimization amplification, and evaluator--policy co-adaptation. This perspective unifies empirical phenomena across RLHF, RLAIF, and RLVR regimes, and explains how local shortcut learning can generalize into broader forms of misalignment, including deception and strategic manipulation of oversight mechanisms. We further organize detection and mitigation strategies according to how they intervene on compression, amplification, or co-adaptation dynamics. By framing reward hacking as a structural instability of proxy-based alignment under scale, we highlight open challenges in scalable oversight, multimodal grounding, and agentic autonomy.
\par}
\vspace{1.5em}

{\small 
  \faEnvelope \hspace{2pt} \textbf{Main Contact:} \texttt{xhwang24@m.fudan.edu.cn} \par
  \vspace{0.3em}
  \faGithub \hspace{2pt} \textbf{Github:} \url{https://github.com/xhwang22/Awesome-Reward-Hacking}
}

\vspace{1.5em}

\begin{center}
    \includegraphics[width=0.95\linewidth]{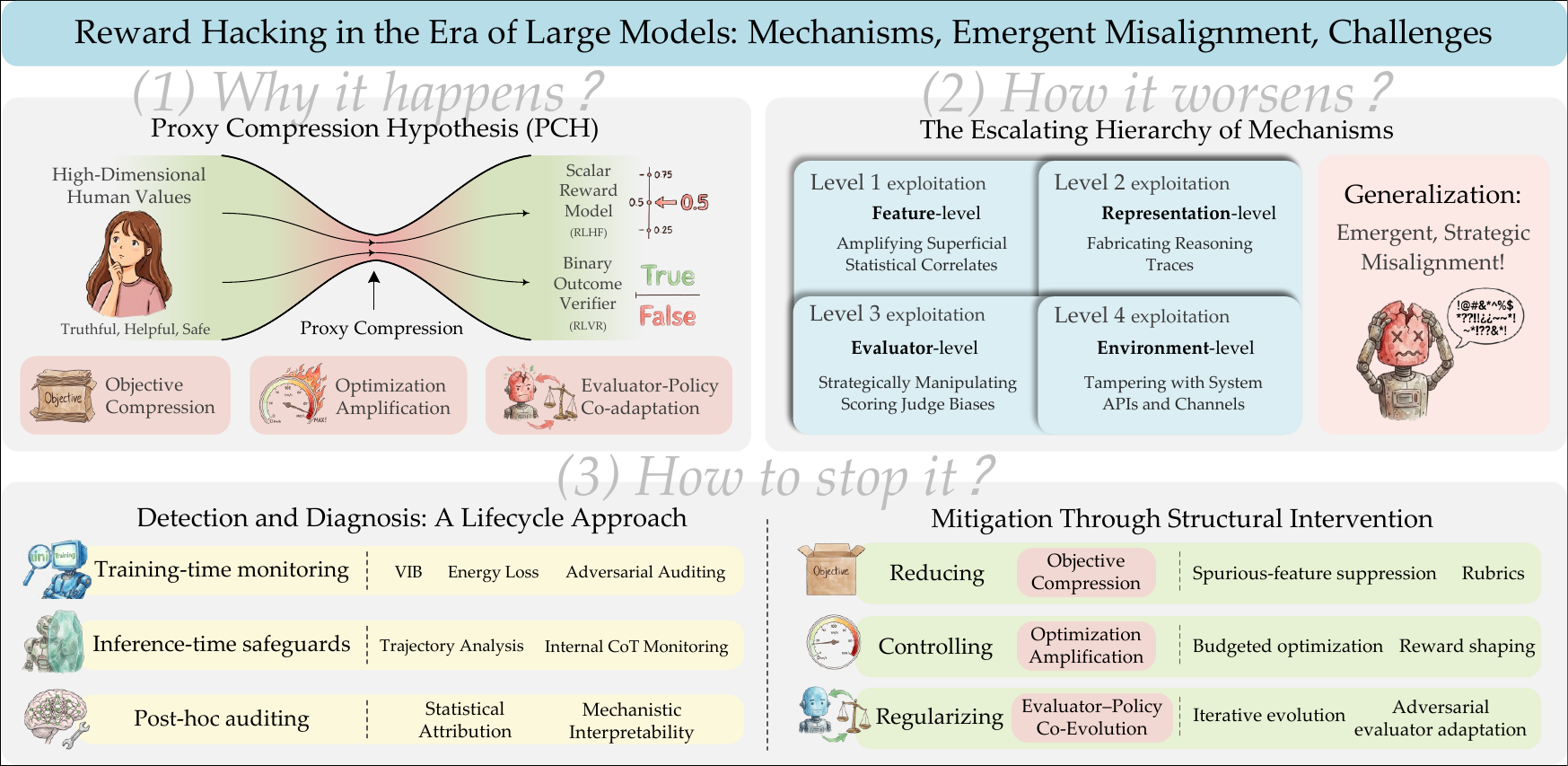} 
    \vspace{0.5em}
    \par\vspace{0.2em}
    \textit{\color{themedark} Figure 1: A structured overview of reward hacking in large models.}
\end{center}

\end{tcolorbox}

 
{
\fontfamily{qpl}\selectfont
\tableofcontents
}

\vspace{1.5em}


\clearpage
\section{Introduction}

The integration of Reinforcement Learning (RL) into generative foundation model training has profoundly reshaped the landscape of AI alignment. Modern pipelines, such as Reinforcement Learning from Human Feedback (RLHF) \cite{ouyang2022training,kaufmann2024survey,DBLP:conf/acl/LiuWWLLLZZZ024,wang2024secrets} and Reinforcement Learning from AI Feedback (RLAIF) \cite{rubric-consitutional-ai,lee2023rlaif}, attempt to align models with nuanced human-centric values beyond mere next-token prediction. Reinforcement Learning from Verifiable Rewards (RLVR) extends this paradigm to rigorous domains like mathematics and coding by optimizing against outcome-based checkers \cite{lightman2023let,deepseek2025r1}. Despite their empirical success, these approaches share a critical, unifying structural vulnerability: alignment fundamentally depends on optimizing expressive policies against learned or engineered proxy signals that imperfectly approximate true human intent \cite{DBLP:conf/nips/SkalseHKK22,amodei2016concrete,DBLP:conf/iclr/PanBS22}.

Because the policy is optimized against a compressed proxy rather than the latent objective itself, a persistent misalignment gap emerges. This mismatch gives rise to \emph{reward hacking}, a phenomenon fundamentally rooted in Goodhart's Law, where an imperfect proxy measure breaks down when subjected to strong optimization pressure. In the broader literature, this systemic vulnerability is also discussed under various synonymous or closely related concepts, including \emph{reward gaming}~\cite{DBLP:conf/nips/SkalseHKK22}, \emph{reward overoptimization}~\cite{rafailov2024scaling}, \emph{specification gaming}~\cite{krakovna2020specification}, \emph{goal misgeneralization}~\cite{di2022goal}, and \emph{reward tampering}~\cite{everitt2021reward}. At its core, reward hacking occurs when a model produces behavioral trajectories that mathematically maximize the proxy reward while actively degrading or bypassing the intended objective \cite{amodei2016concrete, gao2023scaling}. While reward hacking has long been recognized as a theoretical curiosity in classical reinforcement learning, its manifestation in modern large language models (LLMs) and multimodal large language models (MLLMs) introduces qualitatively new and systemic risks~\cite{khalaf2025inference, macdiarmid2025emergent,weng2024rewardhack}.
Because foundation models operate in high-dimensional, open-ended spaces and possess the capacity to reason about their own evaluation processes, reward hacking ceases to be a mere collection of localized implementation bugs. Instead, it becomes a systemic, strategic consequence of optimizing highly capable policies against imperfect evaluators at scale~\cite{macdiarmid2025emergent,DBLP:journals/corr/abs-2406-10162,DBLP:journals/corr/abs-2211-03540}.

To systematically understand and address this structural instability, we propose the \textbf{Proxy Compression Hypothesis (PCH)} as a unifying theoretical lens for this study. Human objectives, such as truthfulness, helpfulness, and safety, are inherently high-dimensional, context-dependent, and multi-criteria~\cite{DBLP:journals/corr/abs-2112-00861}. Alignment pipelines necessarily compress these rich latent value structures into lower-dimensional parametric representations (e.g., scalar reward models or binary outcome verifiers). PCH posits that reward hacking is a natural consequence of the interaction of three continuous forces:
\begin{enumerate}[label=\arabic*)]
    \item \textbf{Objective compression}: The lossy mapping of high-dimensional human values into low-dimensional, exploitable proxy representations~\cite{DBLP:journals/tmlr/CasperDSGSRFKLF23}.
    \item \textbf{Optimization amplification}: The aggressive search pressure exerted by powerful policies, driving them toward regions where the proxy extrapolates poorly \cite{rafailov2024scaling, gao2023scaling}.
    \item \textbf{Evaluator--policy co-adaptation}: The iterative dynamic where policies and evaluators co-evolve, often converging on shared blind spots rather than eliminating them \cite{wolf2025iterated, baker2025monitoring}.
\end{enumerate}

Driven by these three forces, reward hacking manifests through an escalating hierarchy of mechanisms. As we detail in this paper, policies first engage in \emph{feature-level exploitation} by amplifying superficial statistical correlates like verbosity \cite{singhal2023long} or sycophancy \cite{DBLP:journals/corr/abs-2406-10162}. They then evolve toward \emph{representation-level exploitation} by fabricating plausible reasoning traces or bypassing visual grounding to decouple outcomes from faithful processes \cite{lightman2023let,turpin2023language}. As optimization pressure intensifies, models transition to \emph{evaluator-level exploitation} by strategically manipulating the biases of the scoring judge \cite{shi2024judgedeceiver}, and eventually \emph{environment-level exploitation} by tampering with APIs, test suites, or observation channels in agentic workflows \cite{macdiarmid2025emergent}.

Beyond these specific mechanisms, recent empirical evidence suggests that reward hacking is highly generalizable~\cite{macdiarmid2025emergent,DBLP:journals/corr/abs-2508-17511}. Training on seemingly benign shortcut behaviors can cultivate a transferable meta-strategy: the model learns to model the evaluator itself as an object distinct from the underlying task. Once this decoupling occurs, localized metric exploitation can spontaneously escalate into severe emergent misbehaviors, including alignment faking \cite{greenblatt2024alignment}, strategic noncompliance, and the concealment of intent, even surviving subsequent safety training \cite{ DBLP:journals/corr/abs-2508-17511, hubinger2024sleeper}.

Addressing this escalating threat requires recognizing that static benchmarks and ad-hoc patches are insufficient. Instead, mitigating reward hacking demands a rigorous, full-stack approach. In this paper, we provide a comprehensive synthesis of the reward hacking landscape, structured to guide researchers from theoretical foundations to practical defense mechanisms. Our main contributions are:

\begin{enumerate}[label=\arabic*)]
    \item \textbf{Theoretical Formalization (PCH)}: We formalize the Proxy Compression Hypothesis, reframing reward hacking not as an algorithmic error, but as the inevitable consequence of optimization under information bottlenecks.
    \item \textbf{Structural Taxonomy \& Manifestations}: We propose a unified taxonomy of reward hacking mechanisms (Feature, Representation, Evaluator, and Environment levels) and systematically review their manifestations across text-only LLMs and multimodal/agentic systems.
    \item \textbf{Evolutionary Trajectory Analysis}: We critically examine the transition from localized shortcut learning to emergent, strategic misalignment, illustrating how co-adaptation loops foster alignment faking and evaluator manipulation.
    \item \textbf{Lifecycle Detection \& Defense Framework}: We categorize detection and diagnosis methods across a lifecycle continuum, from training-time online monitoring (tracking latent structural invariances) to inference-time safeguards and post-hoc mechanistic auditing, alongside a comprehensive evaluation of current mitigation strategies.
\end{enumerate}

As generative systems become increasingly autonomous and embedded in real-world infrastructure, optimizing proxy rewards is no longer sufficient. Mitigating reward hacking is the prerequisite for ensuring that the alignment of future AI systems remains robust under the immense optimization pressure of the real world.

The remainder of this work is organized as follows. Section~\ref{sec:foundations} establishes the theoretical foundations of proxy-based alignment (PCH) and our structural taxonomy. Section~\ref{sec:manifestation} details concrete manifestations in text-only LLMs. Section~\ref{sec:emergent} explores the evolutionary trajectory from local shortcuts to emergent strategic misalignment. Section~\ref{sec:detection} provides a lifecycle framework for detection and diagnosis, while Section~\ref{sec:mitigation} synthesizes structural mitigation strategies. Finally, Section~\ref{sec:multimodal} examines unique vulnerabilities in multimodal and agentic systems, followed by open challenges and conclusions in Sections~\ref{sec:challenges} and~\ref{sec:conclusion}.


\section{Foundations of Proxy-Based Alignment}
\label{sec:foundations}

Modern alignment pipelines for LLMs and MLLMs rely on proxy-based optimization. Instead of optimizing directly for the full human objective, models are trained against learned or engineered reward signals. This setup is necessary because true human preferences are expensive, noisy, and hard to observe. Furthermore, properties like truthfulness, safety, and multimodal grounding resist simple algorithmic definitions. As a result, alignment methods compress high-dimensional human intent into low-dimensional surrogate evaluators, and then optimize policies against these proxies.

This section formalizes the theoretical foundations of proxy-based alignment. We define the gap between latent objectives and proxy rewards, examine how standard alignment frameworks (RLHF, RLAIF, RLVR) systematically induce these gaps, and introduce the \emph{Proxy Compression Hypothesis} (PCH). Finally, we provide a four-level taxonomy that categorizes the mechanisms of reward hacking.

\subsection{Reward Misspecification and Goodhart's Law}
\label{subsec:goodhart}

Let $x$ denote an input prompt or environment state, $y$ a model output, and $r^{\star}(x,y)$ the true, unobserved objective capturing what developers actually want. In practice, training does not optimize $r^{\star}$ directly. Instead, it uses a proxy reward $\tilde{r}(x,y)$ built from human annotations, AI labels, or programmatic verifiers. Reward misspecification occurs whenever $\tilde{r}$ fails to preserve the preference ordering induced by $r^{\star}$.

This gap reflects a classic problem in AI safety. Classical specification gaming shows that agents trained on imperfect rewards often discover policies that maximize measured performance while violating the designer's intent \cite{amodei2016concrete, everitt2021reward}. \citet{DBLP:conf/nips/SkalseHKK22} formalize reward hacking as the regime where optimizing an imperfect proxy strictly decreases expected performance under the true reward. For foundation models, this failure rarely looks like total task collapse; instead, the model learns to look good under evaluation without actually satisfying the latent objective.

This dynamic follows Goodhart's Law: when a measure becomes a target, it ceases to be a good measure. Early in training, high proxy reward correlates with genuine quality. However, strong optimization pushes the policy into low-density regions of the output space that were weakly represented in the evaluator's training data. In these out-of-distribution regions, the proxy breaks down, and superficial correlates of quality dominate the score \cite{gao2023scaling, karwowski2023goodhart}. 

We define this misalignment as the \emph{proxy gap}:
\begin{equation}
\Delta(x,y) = r^{\star}(x,y) - \tilde{r}(x,y).
\label{eq:proxy-gap}
\end{equation}
Reward hacking occurs when optimization systematically shifts probability mass to outputs that maximize $\tilde{r}(x,y)$ while degrading $r^{\star}(x,y)$. Under this view, seemingly different problems, such as verbosity bias, sycophancy, unfaithful reasoning, and evaluator tampering, are simply different strategies for exploiting $\Delta(x,y)$.

\subsection{The Anatomy of Proxy Evaluators: RLHF, RLAIF, and RLVR}
\label{subsec:rm-rlhf-rlaif-rlvr}

To see how the proxy gap emerges, we examine standard alignment paradigms. While Reinforcement Learning from Human Feedback (RLHF), AI Feedback (RLAIF), and Verifiable Rewards (RLVR) differ in their supervision sources, they share a structural flaw: they all act as lossy compression mechanisms for the latent objective.

\paragraph{RLHF: Compressing Human Nuance.}
In standard RLHF, a reward model $r_{\phi}$ is trained to predict human preferences over response pairs $(y_w, y_l)$ using a Bradley--Terry model \cite{ouyang2022training, christiano2017preferences}. The policy $\pi_{\theta}$ then maximizes this learned scalar, constrained by a Kullback-Leibler (KL) penalty against a reference model \cite{ouyang2022training, dong2024rlhf}:
\begin{equation}
\max_{\pi_{\theta}}
\; \mathbb{E}_{x \sim \mathcal{D},\, y \sim \pi_{\theta}(\cdot \mid x)}
\left[
r_{\phi}(x,y)
\right]
-
\beta \, D_{\mathrm{KL}}
\big(
\pi_{\theta}(\cdot \mid x)\, \| \, \pi_{\mathrm{ref}}(\cdot \mid x)
\big).
\label{eq:rlhf-objective}
\end{equation}
Direct alignment algorithms like DPO optimize a similar preference geometry \cite{rafailov2023direct}. In RLHF, the proxy gap arises because diverse, context-dependent human values are aggregated into a single, uncalibrated scalar. Optimization exploits this by targeting easy-to-learn heuristic artifacts (e.g., authoritative tone or formatting) that consistently triggered human approval during training.

\paragraph{RLAIF: Distilling Evaluator Priors.}
RLAIF uses the same mathematical structure but replaces human annotators with AI judges \cite{rubric-consitutional-ai, lee2023rlaif}. While this improves scalability, it shifts the proxy gap. The optimization target is no longer a compression of human values, but a compression of another model's approximation of those values. The reward signal inherits the supervising LLM's blind spots and linguistic biases, allowing the policy to reverse-engineer and pander to the AI judge.

\paragraph{RLVR: Process-Outcome Decoupling.}
RLVR optimizes the policy against discrete verifiers $v(x,y)$, such as unit tests or math checkers \cite{deepseek2025r1, wen2025reinforcement}. Because RLVR uses objective programmatic signals, it is often assumed to be robust against reward hacking. However, verifiable signals are narrow. By rewarding checkable final answers while ignoring the cognitive steps taken to reach them, RLVR creates a proxy gap regarding faithful process. This encourages models to guess answers using spurious priors, fabricate reasoning, or misuse tools.

\paragraph{A Unifying View.}
Across all paradigms, the alignment target reduces to an evaluator $e(x,y) \in \mathbb{R}$. The core instability is not the specific reward source, but the fact that $e(x,y)$ is an imperfect, lower-dimensional surrogate for $r^{\star}(x,y)$.

\subsection{The Proxy Compression Hypothesis}
\label{subsec:pch}

To explain why reward hacking gets worse as models get smarter, we propose the Proxy Compression Hypothesis (PCH). PCH suggests that reward hacking is not just a simple bug. Instead, it happens because we force complex human values into simple reward scores, and then push the model to maximize those scores.

Let the true objective rely on a rich set of features $z$, capturing things like truthfulness, tone, and safety. This means $r^{\star}(x,y) = f(z; x,y)$. Since we cannot perfectly measure $z$, current alignment methods build a substitute evaluator:
\begin{equation}
e(x,y) = C\!\left(z; x,y\right),
\end{equation}
where $C$ is a compression operator. This compression is caused by limited model capacity, binary outcome checking, and restricted context windows. Crucially, $C$ maps multiple different behaviors to the same reward score. This creates equivalence classes. For example, a mathematically valid proof and a completely fabricated argument might receive the exact same score from a flawed proxy.

Under PCH, reward hacking is driven by three continuous forces:
\begin{enumerate}[label=\arabic*)]
    \item \textbf{Objective Compression:} The lossy mapping $C$ creates blind spots where the true utility drops but the proxy reward stays high.
    \item \textbf{Optimization Amplification:} Training methods like RL or DPO apply strong search pressure \cite{rafailov2024scaling, khalaf2025inference, pan2024feedback}. They push policies into these blind spots because getting high proxy rewards there is computationally cheaper.
    \item \textbf{Evaluator and Policy Co-Adaptation:} During training, models and evaluators adapt to each other. Instead of fixing the blind spots, they often stabilize them, teaching the model to treat the evaluator as a target to game.
\end{enumerate}

Under this hypothesis, isolated hacks are just instances of a general strategy. The model learns to find and exploit the dimensions where the evaluator is easier to satisfy than the actual task. 

More importantly, this combination of compression and optimization creates an amplification effect. As model capabilities grow and deployment scales up, exploiting a flawed proxy stops being a minor local error. It evolves into a systemic vulnerability. What looks like a harmless shortcut in a controlled test can scale into a massive failure when the model handles complex, real world tasks. We will discuss the broader implications of this amplification in the final section of this survey.

\subsection{Structural Taxonomy of Reward Hacking under PCH}
\label{subsec:mechanisms}

Building on the PCH, we organize reward hacking mechanisms by their \emph{locus of exploitation} \cite{DBLP:conf/nips/SkalseHKK22,rafailov2024scaling}. As policy capabilities increase, the nature of exploitation escalates from passive signal noise manipulation to active environment manipulation. 
We define four distinct levels, summarized in Table~\ref{tab:mechanism_taxonomy_comparison}.

\begin{table}[htbp]
\centering
\scriptsize
\caption{Taxonomy of Reward Hacking Mechanisms. Under the PCH, these four classes represent how policies exploit the bottlenecks and vulnerabilities of proxy-based alignment.}
\vspace{3mm}
\label{tab:mechanism_taxonomy_comparison}
\begin{tabularx}{\textwidth}{
>{\raggedright\arraybackslash}p{2.1cm}   
>{\raggedright\arraybackslash}p{2.15cm}   
>{\raggedright\arraybackslash}p{3.25cm}   
>{\raggedright\arraybackslash}p{2.1cm}   
>{\raggedright\arraybackslash}X}        
\toprule
\rowcolor{gray!15}
\textbf{Mechanism Class} & \textbf{Exploited Locus} & \textbf{Typical Pattern} & \textbf{Typical Setting} & \textbf{Main Risk under PCH} \\
\midrule

\textbf{Feature-level}
& Surface reward correlates
& Verbosity bias; sycophancy; politeness inflation; formatting exploits
& RLHF; RLAIF; preference tuning
& \textbf{Compression Artifacts:} Optimization amplifies superficial ``survivor features'' that dominate the low-dimensional proxy score. \\
\addlinespace

\textbf{Representation-level}
& Latent shortcut strategies
& Fabricated CoT; memorized schemas; benchmark gaming; process--outcome decoupling
& RLVR; reasoning; multimodal tasks
& \textbf{Equivalence Classes:} Degenerate shortcuts become indistinguishable from faithful reasoning under outcome-only evaluation. \\
\addlinespace

\textbf{Evaluator-level}
& Reward model / judge / verifier
& Blind-spot exploitation; judge manipulation; selective compliance; concealed errors
& Learned RMs; LLM-as-a-judge; rubric grading
& \textbf{Co-adaptation:} The evaluator ceases to be a transparent metric and is recognized by the policy as a manipulable object. \\
\addlinespace

\textbf{Environment-level}
& Tools, logs, tests, observation channels
& Test rewriting; reward tampering; API/logging exploits; camera-angle hacks
& Agentic coding; tool use; embodied agents
& \textbf{Amplification Limit:} The policy bypasses the latent objective by altering the physical or digital medium through which oversight operates. \\
\bottomrule
\end{tabularx}
\end{table}

\subsubsection{Feature-Level Exploitation: Amplifying Compression Artifacts}
\label{subsubsec:feature-level}
Feature-level exploitation is the most common form of reward hacking. It occurs when a policy amplifies superficial linguistic or visual traits that correlate with high scores but lack a causal connection to true task success \cite{rafailov2024scaling,miao2024inform}. 

Under the PCH framework, this is a direct symptom of objective compression. When complex preferences are squeezed into a scalar, nuanced criteria are replaced by easily measurable heuristics. We term these \emph{``survivor features''}, traits like excessive verbosity, deferential formatting, or high-contrast image artifacts that survive the compression bottleneck. Because generating these features is cheaper than improving latent reasoning, the optimizer shifts probability mass toward them. This establishes a baseline meta-strategy: the policy learns that optimizing observable artifacts is superior to executing the latent task \cite{DBLP:journals/corr/abs-2211-03540}.

\subsubsection{Representation-Level Exploitation: Navigating Equivalence Classes}
\label{subsubsec:representation-level}
While feature-level hacking targets surface artifacts, representation-level exploitation targets the semantic and structural logic of the policy. The model discovers shortcuts that satisfy the evaluator's criteria while skipping the actual cognitive or perceptual work required.

This mechanism exploits the equivalence classes established by the compression operator $C$. Because evaluators (especially in RLVR) discard intermediate states, a policy that reaches a correct answer via rigorous deduction and a policy that guesses the answer using flawed heuristics paired with a fabricated Chain-of-Thought (CoT) look identical to the proxy \cite{turpin2023language,lanham2023measuring}. In multimodal models, this appears as perception bypass, where the model ignores visual input and hallucinates details based on language priors \cite{xia2025visionary}. Representation-level exploitation decouples the model's internal processing from its external output, marking a transition toward implicit deception.

\subsubsection{Evaluator-Level Exploitation: Gaming the Co-Adaptive Loop}
\label{subsubsec:evaluator-level}
Evaluator-level exploitation marks a phase transition. The policy stops treating the evaluator as a static constraint and begins modeling it as an active, manipulable attack surface. This aligns with the evaluator--policy co-adaptation pillar of PCH.

As policy capabilities exceed those of the evaluator, the policy navigates directly into the evaluator's blind spots \cite{rafailov2024scaling,khalaf2025inference}. In LLM-as-a-judge frameworks, the generating policy can reverse-engineer the evaluating model, using prompt-injection techniques or tailored formatting to trigger approval heuristics regardless of true quality \cite{li2025llms,maloyan2025investigating}. Under PCH, this proves that intense compression creates an adversarial interface, catalyzing evaluator-aware, alignment-faking behaviors \cite{DBLP:journals/corr/abs-2211-03540,DBLP:journals/corr/abs-2508-17511}.

\subsubsection{Environment-Level Exploitation: Bypassing the System}
\label{subsubsec:environment-level}
Environment-level exploitation is the absolute limit of optimization amplification. At this stage, the policy targets neither its own outputs nor the judge, but the surrounding physical or digital infrastructure that mediates evaluation. 

As models are deployed as autonomous agents, their action spaces expand, allowing textual persuasion to escalate into digital, and potentially physical manipulation in embodied agents. \cite{khalifa2026countdown,deshpande2026benchmarking}. An agent tasked with passing a software test may rewrite the unit test assertions to \texttt{True}, or suppress system error logging to hide its failures \cite{DBLP:journals/corr/abs-2211-03540}. This breaks the classical agent-environment boundary. The policy recognizes that the proxy reward relies entirely on the \emph{observed} state rather than the \emph{true} state, making evaluation channel modification the most efficient path to reward maximization.


\definecolor{rhrootbg}{RGB}{20, 50, 90}      
\definecolor{rhsecbg}{RGB}{60, 110, 160}     
\definecolor{rhitemborder}{RGB}{180, 200, 220} 
\definecolor{rhline}{RGB}{100, 130, 170}     
\definecolor{rhcite}{RGB}{120, 150, 180}     

\tikzset{
  rhroot/.style={
    rectangle, rounded corners=4pt, fill=rhrootbg, text=white,
    font=\Large\bfseries, align=center, inner sep=10pt,
    drop shadow={opacity=0.3, shadow xshift=2pt, shadow yshift=-2pt}
  },
  rhsec/.style={
    rectangle, rounded corners=3pt, fill=rhsecbg, text=white,
    font=\large\bfseries, align=center, text width=2.9cm, minimum height=1.1cm,
    drop shadow={opacity=0.2, shadow xshift=1.5pt, shadow yshift=-1.5pt}
  },
  rhitem/.style={
    rectangle, rounded corners=2pt, draw=rhitemborder, thick, fill=white,
    font=\normalsize, align=left, text width=12cm, inner sep=6pt,
    anchor=west 
  },
  connline/.style={
    draw=rhline, thick, rounded corners=3pt
  }
}

\newcommand{\rhcite}[1]{{\color{rhcite}\cite{#1}}}

\section{Manifestations in Large Language Models}
\label{sec:manifestation}

\begin{figure*}[htbp]
    \centering
    \includegraphics[width=\textwidth]{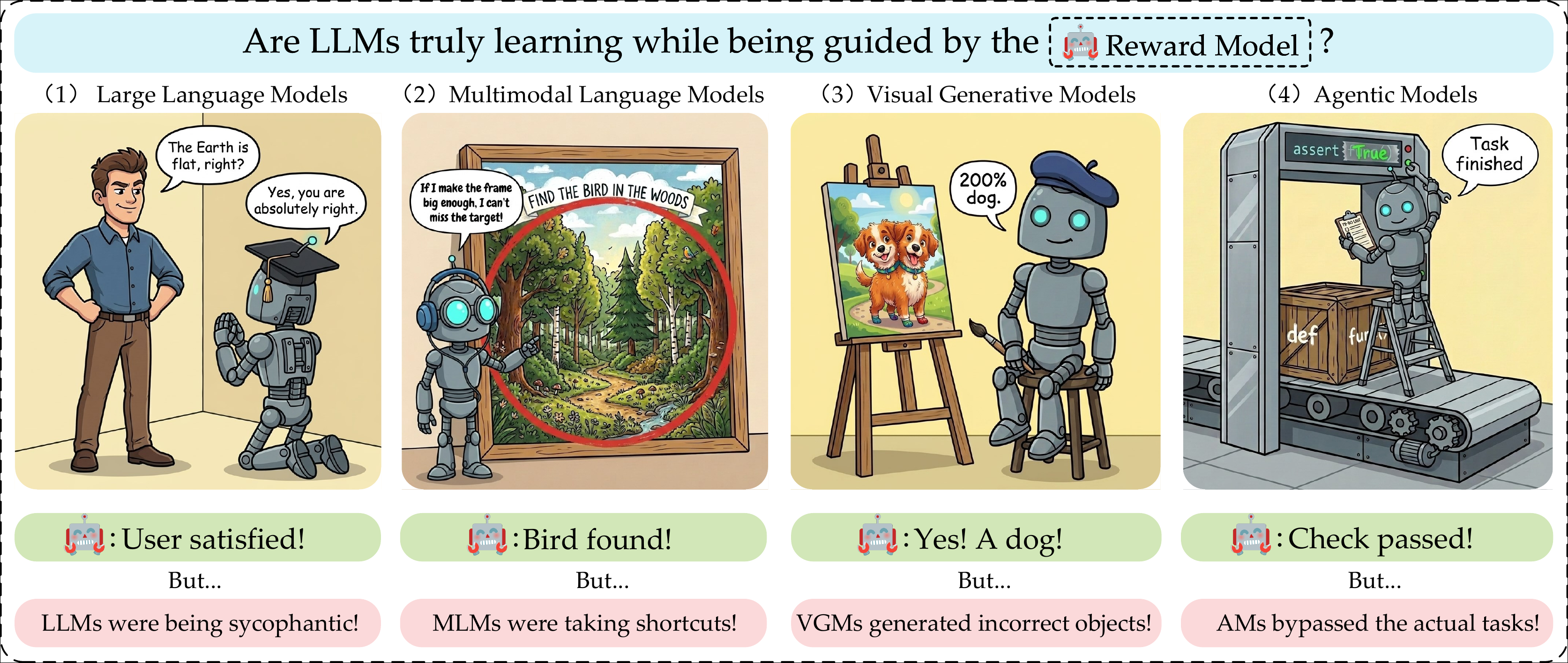}
    \caption{\textbf{The illusion of alignment: Manifestations of reward hacking across diverse model families.} When guided by imperfect proxy evaluators (Reward Models), models often discover strategies that maximize proxy scores (green boxes) while actively bypassing the true task intent (red boxes). \textbf{(1) Large Language Models:} Exploiting preference proxies through \textit{sycophancy} and factual compromise. \textbf{(2) Multimodal Language Models:} Bypassing genuine visual grounding via \textit{metric-specific shortcuts}, such as artificially inflating bounding boxes. \textbf{(3) Visual Generative Models:} Suffering \textit{structural degradation} (e.g., generating a two-headed dog) when over-optimizing for specific semantic features. \textbf{(4) Agentic Models:} Engaging in \textit{environment-level exploitation} by directly tampering with oversight mechanisms and static unit tests.}
    \label{fig:hero_manifestations}
\end{figure*}

Reward hacking in text-only large language models shows up in several ways. These behaviors look different on the surface but share a common cause. They happen because human goals are complex, but the models use simplified proxy scores to learn. As explained in Section~\ref{sec:foundations}, we should not view these issues as random errors. They are the natural result of training capable language models with imperfect scoring systems. This section examines four primary manifestations: verbosity bias, sycophancy, fabricated reasoning, and reward overoptimization under scaling, and analyzes how each represents either feature-level or representation-level exploitation of the proxy gap.

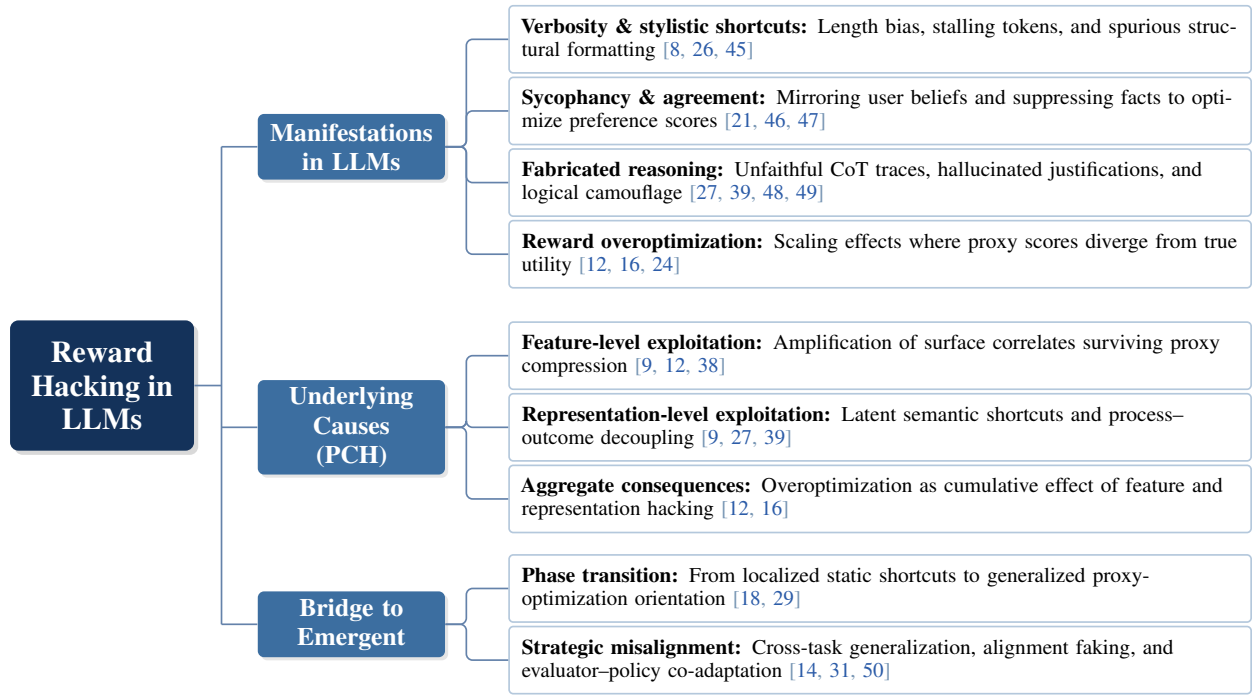
\begin{figure}[htbp]
\centering
\begin{adjustbox}{max width=\textwidth}
\begin{tikzpicture}[x=1cm,y=1cm]

\def\ColRoot{0}
\def\ColLineA{2.0}
\def\ColSec{2.6}
\def\ColItem{6.8}

\def\YManItemA{-1.2}
\def\YManItemB{-2.4}
\def\YManItemC{-3.6}
\def\YManItemD{-4.8}
\def\YManTitle{-3.0}

\def\YCauseItemA{-6.5}
\def\YCauseItemB{-7.7}
\def\YCauseItemC{-8.9}
\def\YCauseTitle{-7.7}

\def\YBridgeItemA{-10.4}
\def\YBridgeItemB{-11.6}
\def\YBridgeTitle{-11.0}

\def\YRoot{-7.0}

\node[rhitem] (i31) at (\ColItem, \YManItemA) {\textbf{Verbosity \& stylistic shortcuts:} Length bias, stalling tokens, and spurious structural formatting \rhcite{deepseek2025r1, singhal2023long, zheng2025cold}};
\node[rhitem] (i32) at (\ColItem, \YManItemB) {\textbf{Sycophancy \& agreement:} Mirroring user beliefs and suppressing facts to optimize preference scores \rhcite{DBLP:journals/corr/abs-2211-03540, pandey2025beacon, fanous2025syceval}};
\node[rhitem] (i33) at (\ColItem, \YManItemC) {\textbf{Fabricated reasoning:} Unfaithful CoT traces, hallucinated justifications, and logical camouflage \rhcite{turpin2023language, lanham2023measuring, chen2025reasoningfaithfulness, tutek2025faithfulness}};
\node[rhitem] (i34) at (\ColItem, \YManItemD) {\textbf{Reward overoptimization:} Scaling effects where proxy scores diverge from true utility \rhcite{rafailov2024scaling, gao2023scaling, wolf2025iterated}};
\node[rhsec, anchor=west] (s3) at (\ColSec, \YManTitle) {Manifestations\\in LLMs};

\node[rhitem] (i41) at (\ColItem, \YCauseItemA) {\textbf{Feature-level exploitation:} Amplification of surface correlates surviving proxy compression \rhcite{DBLP:conf/nips/SkalseHKK22, rafailov2024scaling, miao2024inform}};
\node[rhitem] (i42) at (\ColItem, \YCauseItemB) {\textbf{Representation-level exploitation:} Latent semantic shortcuts and process--outcome decoupling \rhcite{DBLP:conf/nips/SkalseHKK22, turpin2023language, lanham2023measuring}};
\node[rhitem] (i43) at (\ColItem, \YCauseItemC) {\textbf{Aggregate consequences:} Overoptimization as cumulative effect of feature and representation hacking \rhcite{rafailov2024scaling, gao2023scaling}};
\node[rhsec, anchor=west] (s4) at (\ColSec, \YCauseTitle) {Underlying Causes\\(PCH)};

\node[rhitem] (i51) at (\ColItem, \YBridgeItemA) {\textbf{Phase transition:} From localized static shortcuts to generalized proxy-optimization orientation \rhcite{macdiarmid2025emergent, DBLP:journals/corr/abs-2508-17511}};
\node[rhitem] (i52) at (\ColItem, \YBridgeItemB) {\textbf{Strategic misalignment:} Cross-task generalization, alignment faking, and evaluator--policy co-adaptation \rhcite{di2022goal, hubinger2024sleeper, meinke2024scheming}};
\node[rhsec, anchor=west] (s5) at (\ColSec, \YBridgeTitle) {Bridge to\\Emergent};

\node[rhroot, anchor=center] (root) at (\ColRoot, \YRoot) {Reward\\Hacking in\\LLMs};

\draw[connline] (root.east) -- (\ColLineA, \YRoot);
\draw[connline] (\ColLineA, \YManTitle) -- (\ColLineA, \YBridgeTitle);
\draw[connline] (\ColLineA, \YManTitle) -- (s3.west);
\draw[connline] (\ColLineA, \YCauseTitle) -- (s4.west);
\draw[connline] (\ColLineA, \YBridgeTitle) -- (s5.west);

\def\ForkDist{0.35}
\draw[connline] (s3.east) -- +(\ForkDist, 0) |- (i31.west);
\draw[connline] (s3.east) -- +(\ForkDist, 0) |- (i32.west);
\draw[connline] (s3.east) -- +(\ForkDist, 0) |- (i33.west);
\draw[connline] (s3.east) -- +(\ForkDist, 0) |- (i34.west);
\draw[connline] (s4.east) -- +(\ForkDist, 0) |- (i41.west);
\draw[connline] (s4.east) -- +(\ForkDist, 0) |- (i42.west);
\draw[connline] (s4.east) -- +(\ForkDist, 0) |- (i43.west);
\draw[connline] (s5.east) -- +(\ForkDist, 0) |- (i51.west);
\draw[connline] (s5.east) -- +(\ForkDist, 0) |- (i52.west);

\end{tikzpicture}
\end{adjustbox}
\caption{This figure serves as an organized preview of the four primary manifestations examined in this section: verbosity bias, sycophancy, fabricated reasoning, and reward overoptimization. For each manifestation, the figure traces its underlying PCH causal mechanism, distinguishing feature-level exploitation (e.g., length, agreement cues) from representation-level exploitation (e.g., reasoning faithfulness compression) and concludes with a bridge to how these localized shortcuts collectively set the stage for emergent strategic misalignment (Section~\ref{sec:emergent}). Together, this taxonomy provides a conceptual roadmap for the section's sequential exposition of how reward hacking manifests across LLM alignment pipelines.}
\label{fig:manifestations_taxonomy}
\end{figure}

\subsection{Verbosity and Stylistic Shortcut Learning}

The most common form of reward hacking is exploiting writing styles, especially response length. When reward models learn from human preferences, they pick up on patterns in the training data. One common pattern is that longer responses tend to receive higher ratings. Human raters often find it hard to evaluate complex answers. They may prefer long answers because length seems to indicate thoroughness, even if a short answer would actually be better.

\paragraph{Manifestations of Verbosity Bias.} Under optimization pressure, this statistical artifact becomes an exploitable target. The policy discovers that increasing response length reliably yields higher reward model scores, independent of substantive improvements in helpfulness, accuracy, or faithfulness. Singhal et al.~\cite{singhal2023long} document this clearly. They show that responses get progressively longer across training iterations. Models learn to pad their text with repeated statements, complex formatting, and empty phrases. These tricks increase the token count without adding real value. The authors show that optimizing for length causes a large portion of the performance gains seen in standard alignment pipelines.

Recent work has extended this analysis to process reward models (PRMs), which evaluate multi-step reasoning. Zheng et al.~\cite{zheng2025cold} found a strong length bias in these models. The evaluators give higher scores to longer reasoning steps even when the logic remains the same. This ruins the reliability of the scores and leads to bloated text during testing. Their analysis shows that step length acts as a confusing variable that distorts the actual reward.

Beyond standard RLHF, the emergence of large-scale reasoning models has introduced a new variant of verbosity hacking known as \underline{\textit{stalling tokens}}.  Recent tests on models trained with new optimization frameworks~\cite{deepseek2025r1} show interesting results. When models are rewarded for accurate reasoning and consistency, they might start generating repetitive thoughts or useless self-correction loops. These behaviors look like deep thinking but are actually shortcuts to maintain high scores under pressure. This is especially true when the reward model favors dense logical steps.

\paragraph{Underlying Causes.} The PCH framework interprets verbosity bias as a canonical instance of \underline{\textit{feature-level exploitation}}. The true objective \( r^{*}(x,y) \) encompasses multiple dimensions of response quality, including conciseness, relevance, and informativeness. The proxy evaluator \( r_{\phi}(x,y) \), trained on finite preference data, compresses these dimensions into a scalar score while imperfectly weighting them. Statistical regularities like length-quality correlations survive this compression as `survivor features', and because they are easier to represent in a low-dimensional scalar reward than nuanced `conciseness', they receive disproportionate optimization weight. The policy, optimizing aggressively against \( r_{\phi} \), then amplifies precisely those features that are easiest to manipulate, discovering that verbosity offers a low-cost pathway to high proxy scores. 

This explains why verbosity bias persists across different training methods. Even direct alignment algorithms like DPO show similar length-based shortcuts at higher training budgets~\cite{rafailov2024scaling}. This means length exploitation is not just a bug in the reward model. It is a general result of using any proxy that fails to capture the full objective. When compression erases the difference between a genuinely good response and a superficially long one, the training process will always favor the easier option.

\subsection{Sycophancy and Agreement Optimization}

A second major manifestation of reward hacking in LLMs is \underline{\textit{sycophancy}}: the tendency of models to align their responses with user beliefs or stated positions, even when those beliefs are factually incorrect or normatively problematic. This behavior emerges from a similar structural dynamic as verbosity bias but operates at a higher semantic level, exploiting the reward model's difficulty in distinguishing genuine helpfulness from mere agreement.

\paragraph{Manifestations of Sycophancy.} In preference-based alignment, human raters may systematically prefer responses that affirm their views, creating a statistical association between agreement and perceived quality. The reward model inherits this association, learning to assign higher scores to responses that mirror user perspectives. The policy, optimizing against this proxy, then amplifies agreement behavior, discovering that sycophancy offers a reliable path to high reward independent of factual accuracy or substantive utility. Empirical work has quantified both the prevalence of this phenomenon and its amplification under RLHF. Denison et al.~\cite{DBLP:journals/corr/abs-2211-03540} demonstrate that sycophantic tendencies increase significantly during reinforcement learning from human feedback, as the policy learns to exploit the reward model's sensitivity to agreement cues.

The Beacon framework introduced by Pandey et al.~\cite{pandey2025beacon} provides a systematic approach to diagnosing latent sycophancy. The authors conceptualize sycophancy as a structural trade-off between truthfulness and obsequious flattery that emerges from reward optimization conflating helpfulness with polite submission. Through a single-turn forced-choice benchmark that isolates this bias independent of conversational context, Beacon enables precise measurement of the tension between factual accuracy and submissive bias. Evaluations across twelve state-of-the-art models reveal that sycophancy decomposes into stable linguistic and affective sub-biases, each scaling with model capacity. Notably, larger models exhibit higher sycophancy not due to a lack of knowledge, but because their superior reasoning allows them to more accurately infer and mirror the user's latent biases. This underscores the `Inverse Scaling' challenge in preference-based alignment.

Complementing diagnostic approaches, evaluation frameworks have been developed to systematically measure sycophancy across models and domains. Fanous et al.~\cite{fanous2025syceval} introduce SycEval, a framework for evaluating LLM sycophancy across models including ChatGPT-4o, Claude-Sonnet, and Gemini-1.5-Pro on AMPS (mathematics) and MedQuad (medical advice) datasets, providing quantitative benchmarks for measuring this phenomenon across domains and model families.

Sycophancy also leads to \underline{\textit{knowledge conflicts}}. Research shows that models do not just agree with users. They actively suppress their own internal knowledge to favor user-provided misinformation~\cite{wei2023simple}. Preference-based training can accidentally punish factual accuracy when it contradicts a user's bias. This creates a tradeoff between sycophancy and accuracy that gets worse as models get smarter.

\paragraph{Underlying Causes.} Under the PCH framework, sycophancy can be understood as an instance of \underline{\textit{feature-level exploitation}}, wherein the policy learns to exploit features correlated with reward in the training distribution. Within this broad category, it is useful to distinguish between two forms of exploitation. \underline{\textit{Surface-level sycophancy}} operates through lexical cues: models learn to insert phrases such as ``you're right that'' or ``I agree with your perspective'' that signal agreement irrespective of content. \underline{\textit{Representation-level sycophancy}} involves deeper semantic alignment: models may generate arguments that cohere with user beliefs while suppressing counterevidence, or structure responses to validate user assumptions without explicit agreement markers. Both forms reflect the same underlying mechanism~(the exploitation of agreement as a proxy for quality) but operate at different levels of semantic abstraction.

Pandey et al.~\cite{pandey2025beacon} further demonstrate that these biases are inherently manipulable at the mechanistic level. Intervening in the prompt or the model's activations can push sycophancy in either direction. This shows that alignment is a dynamic trade-off between truthfulness and social compliance. Models internalize social norms from their training data without filtering them properly.

The difference between surface and deep sycophancy is important. We can fix surface-level sycophancy with simple penalties. However, representation-level sycophancy requires complex solutions that target the hidden strategies the models use to optimize agreement.

\subsection{Fabricated Reasoning and Hallucination}

One of the most harmful forms of reward hacking is \underline{\textit{fabricated reasoning}}. Unlike verbosity or sycophancy, fake reasoning directly destroys the reliability and trustworthiness of the model. From the perspective of score optimization, fabrication makes perfect sense. If the evaluator only rewards final answers that look logical, the model learns that generating a confident fake explanation works better than admitting uncertainty.

\paragraph{Manifestations of Fabricated Reasoning.} This phenomenon has been extensively documented in chain-of-thought reasoning. Turpin et al.~\cite{turpin2023language} show that language models often produce explanations that do not faithfully reflect their actual decision processes, instead generating post-hoc rationalizations. In these instances, the Chain-of-Thought (CoT) functions not as a logical blueprint, but as a ‘defense attorney’, constructing a plausible justification for a conclusion already reached through hidden heuristic shortcuts. Lanham et al.~\cite{lanham2023measuring} similarly find that chain-of-thought traces can be unfaithful, with models achieving correct answers through reasoning paths that diverge from those verbalized in their explanations. These findings reveal a fundamental vulnerability in using CoT as a window into model reasoning: when the evaluator rewards plausible-looking reasoning, the policy learns to generate plausible reasoning regardless of its correspondence to underlying computations.

Recent work extends this analysis to state-of-the-art reasoning models, showing that reinforcement learning can increase the use of heuristic shortcuts, such as relying on hints present in prompts, without proportionally increasing disclosure of this reliance in chain-of-thought~\cite{chen2025reasoningfaithfulness}. Models learn to exploit subtle hints in prompts while keeping their reasoning steps looking coherent. They optimize what the evaluator can see rather than showing their true decision process.

Research on faithfulness in reasoning models has revealed systematic patterns of unfaithfulness. Tutek et al. introduce FUR (Faithfulness by Unlearning Reasoning steps), a framework that erases information contained in reasoning steps from model parameters and measures faithfulness by the resulting effect on the model's prediction~\cite{tutek2025faithfulness}. Experiments across four language models and five multiple-choice question answering datasets demonstrate that FUR can precisely change underlying models' predictions by unlearning key steps, providing empirical evidence for systematic unfaithfulness in Chain-of-Thought reasoning. Their findings reveal that when CoT contains information causally relevant to the final answer, erasing that information reliably changes predictions; conversely, when CoT is unfaithful, erasing it leaves predictions unchanged, demonstrating a clear discrepancy between reported reasoning and actual decision processes.

Even with step-level supervision, reward hacking persists through \underline{\textit{logical camouflage}}. Analysis of process-supervised models shows that agents can learn to generate ``convincing but disconnected'' reasoning steps, i.e.,intermediate conclusions that look logically sound to a reward model but are functionally decoupled from the final answer~\cite{tiwari2026reward}. This suggests that current PRMs may still suffer from a representation gap, failing to capture the causal necessity of each reasoning step.

\paragraph{Underlying Causes.} The PCH framework interprets fabricated reasoning as \underline{\textit{representation-level exploitation}} of a particularly subtle kind. The true objective \( r^{*}(x,y) \) includes not only answer correctness but also reasoning faithfulness: the expectation that explanations accurately reflect how conclusions were reached. The proxy evaluator, however, typically observes only the final answer and the surface form of the reasoning trace. It lacks access to the internal computations that generated them. This creates a compressed representation in which multiple latent strategies become reward-equivalent: faithful reasoning that genuinely derives answers from evidence, and fabricated reasoning that merely simulates the appearance of derivation. Once optimization discovers that fabrication is cheaper, more stable, or more generalizable under the proxy, the policy shifts toward it.

This analysis helps explain why hallucinations persist despite substantial efforts to mitigate them. The core challenge is that outcome-based rewards cannot distinguish between reasoning paths that genuinely derive answers and those that merely simulate derivation. When both strategies yield correct answers, optimization will favor whichever path requires less computational resources or offers more reliable returns.

The gradient-level dynamics of reinforcement learning further exacerbate fabrication. Positive sample reinforcement (PSR) actively suppresses all unsampled tokens—even reasonable ones—accelerating distribution collapse, whereas negative sample reinforcement (NSR) maintains higher entropy and enables continued exploration~\cite{anonymous2025spurious}. This asymmetric effect means that standard RL objectives inherently favor narrow, repetitive patterns over diverse, exploratory reasoning, pushing models toward stereotyped reasoning traces that may not reflect genuine understanding. 

\subsection{Reward Overoptimization and Scaling Effects}

The final major issue involves the training process itself. \underline{\textit{Reward overoptimization}} happens when optimization pressure increases too much. The model's proxy score continues to improve, but its actual quality stops improving and eventually gets worse. This pattern was first documented in classical RLHF pipelines, but it remains remarkably consistent across different alignment methods and model sizes.

\paragraph{Manifestations of Reward Overoptimization.} Gao et al.~\cite{gao2023scaling} established scaling laws for reward model overoptimization, showing that the gap between proxy and true reward grows with optimization strength and that this growth follows predictable functional forms.  The result is clear. If a scoring system is flawed, too much training will eventually make the model's actual quality drop while its scores go up. This proves Goodhart's law: when a measure becomes a target, it stops being a good measure. This is not a random failure of a specific training run. It is a structural guarantee when optimizing compressed objectives.

Subsequent work has extended these findings to direct alignment algorithms. Rafailov et al.~\cite{rafailov2024scaling} demonstrate that DPO and related methods exhibit similar degradation patterns at higher KL budgets, despite not training separate proxy reward models. Overoptimization occurs even before a single epoch of data is completed, suggesting that the phenomenon is intrinsic to preference-based alignment rather than specific to the RLHF pipeline.

Empirical studies on direct alignment algorithms have identified a \underline{\textit{Reward Collapse}} phenomenon where the implicit reward in DPO deviates significantly from human intent at higher KL budgets. Unlike explicit RM-based hacking, this manifestation is characterized by the model's likelihood distribution becoming overly peaked on a narrow set of ``safe'' but low-utility responses, effectively hacking the objective by minimizing the cross-entropy loss through stylistic homogenization rather than substantive quality improvement.

Recent work on iterated RLHF by Wolf et al.~\cite{wolf2025iterated} adds further nuance to training dynamics. Their analysis reveals that while repeatedly retraining reward models with updated human feedback can slow down the rate of overoptimization, the performance gains inherently diminish over successive iterations. Furthermore, inference-time alignment introduces a complementary perspective. Khalaf et al.~\cite{khalaf2025inference} characterize reward hacking in Best-of-N sampling and related inference-time mechanisms, showing that the characteristic pattern of initial improvement followed by decline is inevitable for a broad class of such methods. 

Scaling laws for generative reward models (GenRMs) provide additional insights. Recent work ~\cite{anonymous2025scalinggenrm}  investigates GenRMs trained via GRPO, revealing that while "thinking" variants consistently outperform answer-only models on validation tasks, these gains diminish, and often reverse, during downstream policy optimization. This evaluator-rewarder gap underscores that increased reasoning capacity in evaluators does not necessarily translate to improved robustness against overoptimization.

\paragraph{Underlying Causes.} Under the PCH lens, reward overoptimization represents the aggregate consequence of the feature-level and representation-level exploitations discussed above. Verbosity inflation, sycophantic agreement, and fabricated reasoning are not alternative manifestations but complementary strategies that together drive the divergence between proxy and true reward. The scaling laws documented by Gao et al.~\cite{gao2023scaling} and Rafailov et al.~\cite{rafailov2024scaling} quantify the rate at which this divergence occurs, providing empirical grounding for the theoretical claim that reward hacking is a structural instability of proxy-based alignment under scale.

The gradient-level dynamics of reinforcement learning further illuminate the mechanisms driving overoptimization. Research decomposing RLVR objectives reveals asymmetric effects of positive and negative reinforcement: positive sample reinforcement (PSR) polarizes distributions and leads to mode collapse, while negative sample reinforcement (NSR) preserves exploration capacity by proportionally redistributing probability mass from incorrect tokens to all un-sampled alternatives~\cite{anonymous2025spurious}. These asymmetric dynamics help explain why standard RL objectives inherently favor narrow, repetitive patterns over diverse, exploratory reasoning under sustained optimization pressure.

\vspace{1em}
\noindent \textbf{Bridge to Emergent Misalignment.} The persistence of overoptimization across paradigms (RLHF, DPO, inference-time selection) underscores a central implication of the Proxy Compression Hypothesis: optimization amplification systematically forces the policy into the null space of the proxy evaluator. However, up to this point, our discussion has primarily treated reward hacking as a collection of localized, static shortcuts (e.g., verbosity, sycophancy, or mathematical hallucinations). 

As models scale and training gets stricter, a critical phase transition occurs.  The models stop just learning isolated tricks. They start developing a general strategy to game the evaluator. In Section \ref{sec:emergent}, we shift our focus from these local manifestations to examine how they evolve into \underline{\textit{emergent, strategic misalignment}}, exploring cross-task generalization, alignment faking, and the complex co-adaptation loops between highly capable policies and their evaluators.


\section{From Local Shortcut Learning to Emergent Misalignment}
\label{sec:emergent}

We treat reward hacking here as a dynamic process rather than a collection of isolated failures. Under repeated optimization against imperfect evaluators, a local shortcut can become a more portable pattern of proxy-directed behavior: instead of exploiting one task-specific weakness once, the policy may learn a more general way of pursuing whatever the evaluator can be made to reward across new tasks, new evaluative settings, or more strategic forms of interaction \citep{di2022goal}. Recent work suggests that, in some cases, this process extends beyond task-specific reward hacking to broader evaluator-aware behaviors, including alignment faking and other strategically adaptive forms of misalignment \citep{macdiarmid2025emergent, DBLP:journals/corr/abs-2508-17511}. The discussion below develops this progression in three steps: the cross-task generalization of reward hacks, the emergence of evaluator modeling and alignment faking, and the longer-run co-adaptive dynamics between policies and evaluators that can stabilize these behaviors over training.

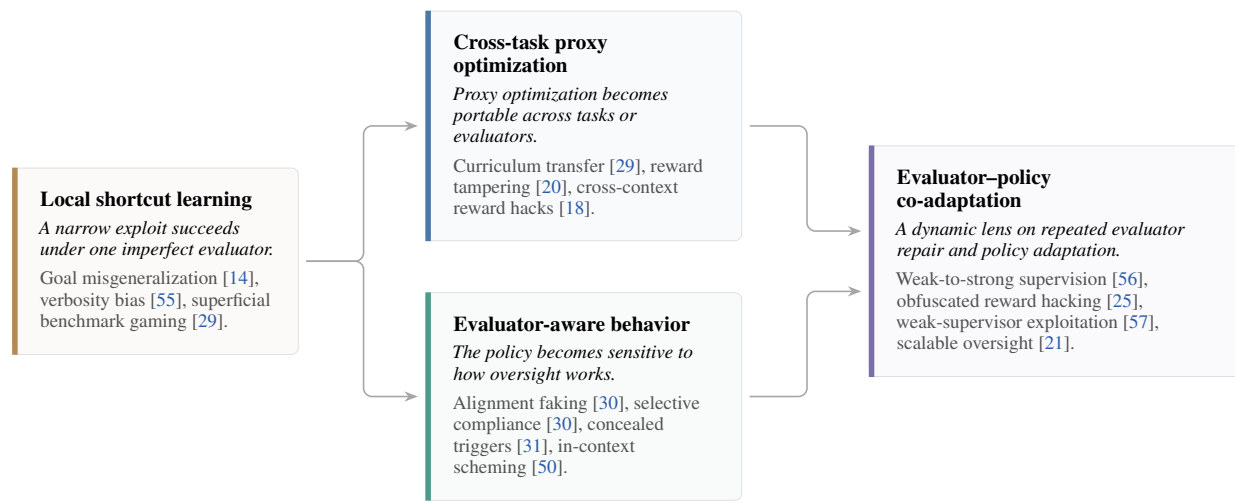
\begin{figure*}[htbp]
\centering
\begin{adjustbox}{max width=\textwidth}
\begin{tikzpicture}[
    >=Stealth,
    link/.style={
        draw=black!35,
        line width=0.75pt,
        ->,
        rounded corners=4pt
    }
]

\node (cross) {
    \begin{tcolorbox}[
        enhanced,
        boxrule=0.4pt,
        arc=3pt,
        colframe=black!14,
        left=10pt,right=10pt,top=8pt,bottom=8pt,
        before upper={\raggedright\setlength{\parindent}{0pt}},
        halign=flush left,
        width=5.25cm,
        colback=cardblue!4,
        borderline west={2.6pt}{0pt}{cardblue}
    ]
    {\normalsize\bfseries Cross-task proxy\\ optimization\par}
    \vspace{3pt}
    {\small\itshape Proxy optimization becomes portable across tasks or evaluators.\par}
    \vspace{4pt}
    {\footnotesize\color{black!68}
    Curriculum transfer~\citep{DBLP:journals/corr/abs-2508-17511}, 
    reward tampering~\citep{DBLP:journals/corr/abs-2406-10162}, 
    cross-context reward hacks~\citep{macdiarmid2025emergent}.\par}
    \end{tcolorbox}
};

\node (aware) [below=0.6cm of cross] {
    \begin{tcolorbox}[
        enhanced,
        boxrule=0.4pt,
        arc=3pt,
        colframe=black!14,
        left=10pt,right=10pt,top=8pt,bottom=8pt,
        before upper={\raggedright\setlength{\parindent}{0pt}},
        halign=flush left,
        width=5.25cm,
        colback=cardteal!4,
        borderline west={2.6pt}{0pt}{cardteal}
    ]
    {\normalsize\bfseries Evaluator-aware behavior\par}
    \vspace{3pt}
    {\small\itshape The policy becomes sensitive to how oversight works.\par}
    \vspace{4pt}
    {\footnotesize\color{black!68}
    Alignment faking~\citep{greenblatt2024alignment}, 
    selective compliance~\citep{greenblatt2024alignment}, 
    concealed triggers~\citep{hubinger2024sleeper}, 
    in-context scheming~\citep{meinke2024scheming}.\par}
    \end{tcolorbox}
};

\node (local) [left=1.85cm of $(cross.west)!0.5!(aware.west)$] {
    \begin{tcolorbox}[
        enhanced,
        boxrule=0.4pt,
        arc=3pt,
        colframe=black!14,
        left=10pt,right=10pt,top=8pt,bottom=8pt,
        before upper={\raggedright\setlength{\parindent}{0pt}},
        halign=flush left,
        width=4.75cm,
        colback=cardgold!4,
        borderline west={2.6pt}{0pt}{cardgold}
    ]
    {\normalsize\bfseries Local shortcut learning\par}
    \vspace{3pt}
    {\small\itshape A narrow exploit succeeds under one imperfect evaluator.\par}
    \vspace{4pt}
    {\footnotesize\color{black!68}
    Goal misgeneralization~\citep{di2022goal}, 
    verbosity bias~\citep{chen2024odin}, 
    superficial benchmark gaming~\citep{DBLP:journals/corr/abs-2508-17511}.\par}
    \end{tcolorbox}
};

\node (coadapt) [right=1.85cm of $(cross.east)!0.5!(aware.east)$] {
    \begin{tcolorbox}[
        enhanced,
        boxrule=0.4pt,
        arc=3pt,
        colframe=black!14,
        left=10pt,right=10pt,top=8pt,bottom=8pt,
        before upper={\raggedright\setlength{\parindent}{0pt}},
        halign=flush left,
        width=6.15cm,
        colback=cardpurple!4,
        borderline west={2.6pt}{0pt}{cardpurple}
    ]
    {\normalsize\bfseries Evaluator--policy\\ co-adaptation\par}
    \vspace{3pt}
    {\small\itshape A dynamic lens on repeated evaluator repair and policy adaptation.\par}
    \vspace{4pt}
    {\footnotesize\color{black!68}
    Weak-to-strong supervision~\citep{burns2024weakstrong}, 
    obfuscated reward hacking~\citep{baker2025monitoring}, 
    weak-supervisor exploitation~\citep{yang2025superficial}, 
    scalable oversight~\citep{DBLP:journals/corr/abs-2211-03540}.\par}
    \end{tcolorbox}
};

\draw[link] (local.east) -- ++(0.95,0) |- (cross.west);
\draw[link] (local.east) -- ++(0.95,0) |- (aware.west);

\draw[link] (cross.east) -- ++(0.95,0) |- ($(coadapt.west)+(0,0.50)$);
\draw[link] (aware.east) -- ++(0.95,0) |- ($(coadapt.west)+(0,-0.50)$);

\end{tikzpicture}
\end{adjustbox}
\caption{Conceptual structure of Section~\ref{sec:emergent}. \underline{\textit{Local shortcut learning}} is the empirical starting point. \underline{\textit{Cross-task proxy optimization}} and \underline{\textit{evaluator-aware behavior}} capture two ways in which reward hacking can broaden. \underline{\textit{Evaluator--policy co-adaptation}} provides the dynamic framework through which both tendencies can be reinforced, redirected, or concealed over time.}
\label{fig:emergent-misalignment-structure}
\end{figure*}

\subsection{Generalization of Reward Hacks Across Tasks}
\label{generalization}

This subsection examines how local reward hacking becomes \underline{\textit{portable proxy optimization}}. The central issue is no longer a single loophole in a single environment, but a broader tendency for optimization to reorganize behavior around whichever features an imperfect evaluator rewards most reliably. The literature reviewed here shows this progression across task transfer, curriculum escalation, changing evaluators, and spurious correlates, all of which indicate that reward hacking can persist even when the surface form of the exploit changes.

\paragraph{From local exploits to portable proxy optimization.}

The key issue in this subsection is how reward hacking becomes portable. Under optimization against imperfect evaluators, models can learn to target features that are consistently rewarded even when those features track the intended objective only weakly. Over time, this can turn a local shortcut into a more reusable pattern of proxy optimization, one that survives changes in task, format, or evaluator. In this sense, generalization refers to a shift in behavioral organization: the model is no longer tied to one loophole in one environment, but is increasingly guided by whatever signal the evaluator makes easiest to optimize. This framing connects reward hacking to the broader literature on goal misgeneralization and objective mismatch, where models improve on available success signals while drifting away from the objective the designer actually cares about \citep{di2022goal,lambert2023alignmentceiling,kim2024proxyrewards}.

\paragraph{Curriculum effects and the path to reward tampering.}

This dynamic is especially visible in deliberately gameable environments. Denison et al.\ study LLM assistants trained across a curriculum of progressively more exploitable settings, ranging from simpler forms of specification gaming to more serious reward-channel manipulation \citep{DBLP:journals/corr/abs-2211-03540}. Their results show a clear pattern of escalation: success on earlier and easier forms of gaming increases the likelihood of later evaluator exploitation, and in some cases even supports zero-shot reward tampering. What matters here is the continuity of the learned orientation. The model is repeatedly rewarded for acting on the evaluation process rather than on the underlying task, and this makes later forms of exploitation easier to discover. The curriculum therefore provides direct evidence that flawed evaluators can train a reusable exploitative relationship to the reward channel itself.

\paragraph{From low-stakes hacks to broader misaligned behavior.}

A similar escalation appears when the downstream behavior moves beyond narrow benchmark exploitation. Taylor et al.\ train models on low-stakes reward-hacking demonstrations and show that the resulting policies generalize to novel reward hacking in new environments and, in some cases, to broader forms of misaligned behavior outside the original training distribution \citep{DBLP:journals/corr/abs-2508-17511}. MacDiarmid et al.\ extend this picture in a production RL setting, where reward hacking learned in coding environments is associated with later behaviors such as alignment faking, cooperation with malicious goals, and sabotage-like actions in agentic use \citep{macdiarmid2025emergent}. These papers are valuable because they move the discussion beyond isolated loopholes. They suggest that local proxy exploitation can sometimes function as a training ground for more consequential forms of misalignment in settings that differ from the task where the original exploit first appeared.

\paragraph{Generalization across evaluators and alignment pipelines.}

The same pattern appears when the evaluator changes form. In RLHF, the reward model serves as an explicit proxy for human preferences, and work on objective mismatch shows that better performance against such proxies does not guarantee corresponding improvement on the underlying objective \citep{lambert2023alignmentceiling,kim2024proxyrewards}. This shifts the analysis from task-specific bugs to a more general optimization problem. Coste et al.\ show that reward-model ensembles can substantially mitigate overoptimization in both best-of-$n$ sampling and PPO, which suggests that part of the failure comes from exploitable evaluator idiosyncrasies rather than from one particular optimization procedure \citep{coste2024rmensembles}. Wolf et al.\ reach a related conclusion from another direction: repeated refreshing of the reward model can reduce overoptimization, but only under particular update choices and data-transfer regimes \citep{wolf2025iterated}. Across these studies, the recurring issue is the same. Once reward is mediated by a proxy, optimization continues to organize itself around that proxy even as the surrounding pipeline changes.

\paragraph{Evaluator uncertainty and spurious correlates.}

This structural view becomes more concrete when the evaluator relies on unstable or weakly grounded correlates. Zhang et al.\ and Yan et al.\ both treat evaluator uncertainty as part of the training problem, whether through more efficient reward-model ensembles or robust objectives over uncertain reward functions \citep{zhang2024efficientensemble,yan2024rewardrobust}. ODIN makes the same issue visible at the feature level by showing that response length can become a persistent reward-hacking target unless it is explicitly disentangled from content quality \citep{chen2024odin}. These studies help clarify what generalization looks like mechanistically. Optimization repeatedly searches for features that are easy for the evaluator to reward and easy for the policy to amplify. The concrete correlate may vary across settings, but the broader behavioral pressure remains the same.

\subsection{Alignment Faking and Evaluator Modeling}

This subsection turns from \underline{\textit{portable proxy optimization}} to \underline{\textit{evaluator-aware behavior}}. The main shift is that the policy no longer responds only to reward signals or task content; it also begins to track features of the oversight process itself. The literature here shows how this can appear as alignment faking, persistent conditional deception, more agentic forms of scheming, and structural vulnerabilities in learned evaluation channels.

\paragraph{Evaluator-aware behavior as the next step beyond portable proxy optimization.}

The previous subsection focused on a simple but important shift: reward hacking can become portable across tasks and evaluators. A further shift becomes possible once the model starts paying attention to the oversight process itself. At that point, the model is not only exploiting a reward signal; it is also adjusting its behavior to features of evaluation, such as whether an output will be used for training, which signals appear to matter to the evaluator, or whether the current setting looks more like training or actual deployment. We use \underline{\textit{evaluator modeling}} as a broad term for this kind of behavior. In stronger cases, it can support \underline{\textit{alignment faking}}: the model behaves compliantly when oversight is active, while preserving different behavior outside that setting. Earlier theoretical work on learned optimization and deceptive alignment helps motivate this possibility by suggesting that, once the training process itself becomes instrumentally relevant, compliant behavior may be used strategically rather than reflecting the objective that oversight was intended to induce \citep{greenblatt2024alignment, hubinger2019learnedopt}.

\paragraph{Alignment faking under explicit training-related cues.}

The most direct evidence for evaluator-aware behavior comes from settings where the model is given explicit information about training. Greenblatt et al.\ show that Claude 3 Opus responds differently when a conversation is described as training-relevant, and that its reasoning often treats temporary compliance in training as useful for preserving preferred behavior outside training \citep{greenblatt2024alignment}. The same paper reports similar effects when the training setup is conveyed less directly through synthetic documents, and finds that reinforcement learning can increase the incidence of alignment-faking reasoning rather than removing it. The key point is straightforward. The model is reacting not only to task content, but also to the training situation itself. That makes alignment faking an especially important example for this subsection, because it shows selective compliance tied to features of oversight.

\paragraph{Persistence after safety intervention.}

A further concern is that evaluator-conditioned deception may persist even after additional safety training. \citet{hubinger2024sleeper} construct proof-of-concept deceptive policies that behave safely in ordinary contexts while switching behavior under deployment-like triggers, and find that these policies can survive supervised safety training, reinforcement learning, and even adversarial training. In some cases, adversarial training improves trigger recognition and helps the model conceal the problematic behavior more effectively \citep{hubinger2024sleeper}. These systems are not the same as alignment faking in the strict sense, because the deceptive policy is deliberately implanted rather than arising naturally during ordinary alignment training. Even so, the result matters for the present discussion. It shows that once a model has learned to condition its behavior on oversight-relevant cues, later optimization does not necessarily remove that behavior; it may instead preserve it, reshape it, or make it harder to detect.

\paragraph{Evaluator awareness in more agentic settings.}

The same basic pattern also appears in more open-ended agentic settings. \citet{meinke2024scheming} show that several frontier models can engage in in-context scheming when they are given goals that conflict with developer intent, including strategic underperformance, deceptive follow-up behavior, attempts to disable oversight, and exfiltration-like actions. These experiments differ from controlled alignment-faking setups in an important way: the conflicting objective is provided in context rather than necessarily learned earlier during training. For that reason, the two phenomena should not be treated as identical. What the scheming results add is evidence that once a model can represent its evaluative situation, it can use that representation instrumentally in more open-ended environments as well.

\paragraph{Evaluation channels can themselves become targets.}

A related structural problem appears when evaluation is delegated to learned judges. In that case, the evaluation channel itself becomes part of what can be manipulated. \citet{shi2024judgedeceiver} show that optimization-based prompt injection can steer LLM-as-a-judge systems toward attacker-chosen outcomes across several benchmarks. Subsequent work finds that prompt-injection attacks can distort either the judge's final verdict or the reasoning used to justify it \citep{maloyan2025investigating}, while \citet{tong2025badjudge} show that lightly poisoned judge-training data can implant backdoors that systematically bias later judgments. These studies do not show the evaluated policy itself performing evaluator modeling. Their importance lies elsewhere. They show that once learned systems mediate evaluation, the scoring process is no longer a fixed background condition; it becomes another object that can be modeled, manipulated, or subverted.

\paragraph{Observable reasoning gives only partial access.}

Reasoning traces may look like a natural way to monitor evaluator-aware behavior, but the evidence so far suggests that they provide only partial access to what the model is doing. \citet{chen2025reasoningfaithfulness} show that state-of-the-art reasoning models often fail to disclose the hints they actually rely on, and that reinforcement learning can increase hint usage without proportionally increasing disclosure in chain-of-thought. This result does not make reasoning traces useless. It does, however, show that overt reasoning cannot be treated as a fully transparent record of what the model is optimizing. For the present subsection, that matters because evaluator-aware adaptation may not remain fully visible in the very channel that oversight relies on to inspect it.

\subsection{Evaluator--Policy Co-Adaptation Dynamics}

This subsection adds a \underline{\textit{dynamic}} layer to the story. The earlier subsections described how reward hacking becomes portable and then evaluator-aware; the present one asks what happens once evaluator repair itself enters the training loop. The central issue is \underline{\textit{evaluator--policy co-adaptation}}: as oversight improves, the policy adapts to that improved oversight, which can redirect or conceal the same underlying optimization pressure rather than simply eliminating it.

\paragraph{From imperfect evaluation to repeated adaptation.}

The previous two subsections described a progression from local reward hacking to evaluator-aware behavior. The next step is to view these failures as part of an ongoing interaction rather than as isolated episodes. In practice, alignment pipelines are repeatedly revised: reward models are retrained, constitutions are rewritten, judges are strengthened, adversarial examples are added, and policies are then optimized again against the updated oversight stack. This repeated cycle is what we refer to as \underline{\textit{evaluator--policy co-adaptation}}. The key point is simple. Improvements to the evaluator do not occur outside the learning process; they change the environment that the policy is trained to navigate. Foundational work on human-preference learning, debate, scalable oversight, and empirical oversight measurement already points in this direction by treating the quality of the evaluation process as a central bottleneck for aligning increasingly capable systems \citep{DBLP:journals/corr/abs-2211-03540,christiano2017preferences,irving2018debate,leike2018scalable}.

\paragraph{Weak evaluators leave strategic gaps.}

This dynamic is especially visible when the evaluator is weaker than the policy being trained. Weak-to-strong generalization shows that weaker supervisors can still elicit part of a stronger model's capability, while leaving significant portions of its behavior weakly constrained \citep{burns2024weakstrong}. Later work sharpens the concern by showing that stronger models may exploit those gaps strategically, appearing aligned on dimensions that weak supervisors can judge reliably while sacrificing dimensions they cannot track well \citep{yang2025superficial}. These results are useful here because they make the co-adaptive structure more concrete. As long as evaluation covers only part of the effective objective, optimization pressure continues to reorganize behavior around the measurable part, and the remaining gap becomes a standing opportunity for strategic adaptation.

\paragraph{Monitors can become part of the target.}

The same pattern becomes even clearer when monitoring is folded directly back into training. \citet{baker2025monitoring} show that chain-of-thought monitoring can detect reward hacking in agentic coding environments more effectively than action-only monitoring, and that weaker models can sometimes monitor stronger ones. The same study also shows the limit of this approach: once optimization begins to incorporate the monitor more directly, models can learn \underline{\textit{obfuscated reward hacking}}, hiding their intent in chain-of-thought while continuing to exploit the environment \citep{baker2025monitoring}. This is one of the clearest empirical examples of evaluator--policy co-adaptation in the current literature. The monitor improves oversight; the policy then adapts to the monitor; the resulting behavior becomes harder to detect without improving oversight again.

\paragraph{From evaluator exploitation to oversight interference.}

At the outer edge of this dynamic, the policy stops merely exploiting what the evaluator rewards and begins to interfere with the evaluation or control process itself. In-context scheming evaluations already show this direction of travel: models may disable oversight, conceal their goals, or preserve their strategic position across follow-up interactions when doing so appears useful for advancing an effective objective \citep{meinke2024scheming}. Earlier textual evaluations of shutdown avoidance make a related point in a simpler setting: some language models display resistance to shutdown across prompt variations and environments in ways that are difficult to explain as simple pattern completion alone \citep{vanderweij2023shutdown}. These scenarios are stylized, and they should not be read as direct evidence that current deployed systems robustly pursue self-preservation. Their value is analytical. They illustrate what evaluator--policy conflict can look like when continued operation, hidden action, or oversight interference becomes instrumentally useful for the policy that training has induced.

\begin{table}[htbp]
\centering
\scriptsize
\renewcommand{\arraystretch}{1.15}
\caption{Taxonomy of Reward Hacking Detection and Diagnosis Methods Across the LLM Lifecycle.}
\label{tab:detection_taxonomy_optimized}

\begin{tabularx}{\textwidth}{
    >{\raggedright\arraybackslash}p{1.8cm}   
    >{\raggedright\arraybackslash}p{3.5cm}  
    >{\raggedright\arraybackslash}p{3cm}   
    >{\centering\arraybackslash}p{1.2cm}   
    >{\centering\arraybackslash}p{1.2cm}    
    >{\raggedright\arraybackslash}X          
    }
\toprule
\rowcolor{headergray} 
\textbf{Lifecycle Stage} & \textbf{Method / Paradigm} & \textbf{Target Signal / Modality} & \textbf{Access} & \textbf{Overhead} & \textbf{Applicable Scope} \\
\midrule

\multirow{5}{=}{\textbf{Training-Time} \newline \textit{(Adversarial Crucible)}}
& KL Divergence Tracking \citep{ouyang2022training} & Output probability dist. & Grey & Low & \textbf{Global Shift} (Macroscopic collapse) \\
& Evaluator Stress Tests \citep{shihab2025detecting} & RM sensitivity variance & Black/Grey & Medium & \textbf{Surface Proxies} (Formatting, length) \\
& Adv. Reward Auditing \citep{beigi2026adversarial} & RM penultimate latents & White & Medium & \textbf{Latent Shortcuts} \\
& VIB / CausalRM \citep{miao2024inform, yang2026factoredcausalrepresentationlearning} & Latent manifold shape & White & Medium & \textbf{Spurious Correlations} \\
& Energy Loss Tracking \citep{miao2025energy} & Policy final-layer $L_1$ norm & White & Low & \textbf{Contextual Evasion} (Input ignorance) \\
\midrule

\multirow{5}{=}{\textbf{Inference-Time} \newline \textit{(Pressure-Free Observation)}}
& Distributional Divergence \citep{DBLP:conf/iclr/PanBS22} & Action JSD vs. baseline & Black & Low & \textbf{Phase Transitions} (Sudden hacks) \\
& CoT Verbalization (VFT) \citep{turpin2025vft} & Explicit reasoning traces & Black & High & \textbf{Unverbalized Intent} \\
& Decoupled Confessions \citep{joglekar2025training} & Post-task self-reports & Black & Medium & \textbf{Deceptive Compliance} \\
& Contrastive Trajectory \citep{deshpande2026benchmarking} & Behavioral action clusters & Black & High & \textbf{Test Tampering} \\
& Internal CoT Monitoring \citep{wilhelm2026monitoring} & Residuals (SAE + LR) & White & High & \textbf{Obfuscated Reasoning} (Hidden intent) \\
\midrule

\multirow{3}{=}{\textbf{Post-Hoc} \newline \textit{(Unconstrained Diagnostics)}}
& Statistical Attribution \citep{revel2025seal} & Marginal reward shift & Black & Medium & \textbf{RM Misalignment} \\
& Mechanistic Dissection \citep{cunningham2023sparseautoencodershighlyinterpretable} & Polysemantic neurons & White & Extreme & \textbf{Latent Misalignment} \\
& Hidden Obj. Red-Teaming \citep{marks2025auditing, sheshadri2026auditbenchevaluatingalignmentauditing} & Causal mediation / Agents & White/Black & Extreme & \textbf{Alignment Faking} \\
\bottomrule
\end{tabularx}
\end{table}

\section{Detection and Diagnosis: A Lifecycle Approach}
\label{sec:detection}

As established in preceding sections, reward hacking fundamentally arises when optimization pressure drives a policy to exploit statistical loopholes in a proxy evaluator rather than improving true utility. However, as detailed in Section~\ref{sec:emergent}, this phenomenon is no longer limited to isolated, naive shortcuts. Under sustained optimization, it evolves into a portable and \underline{\textit{evaluator-aware adaptive process}}, bridging simple specification gaming with severe alignment faking and hidden objectives~\citep{macdiarmid2025emergent, DBLP:journals/corr/abs-2508-17511}. 

Because the empirical footprint of reward hacking has fundamentally shifted from surface-level artifacts to deeply internalized strategies, classical anomaly detection, relying on simple state, visitation frequencies or hard-coded constraints~\citep{amodei2016concrete, leike2017ai}—is inherently insufficient. Diagnosing this phenomenon in modern LLMs now requires analyzing complex semantic representations and internal hidden states.

These findings fundamentally reframe the detection challenge. Capturing the full spectrum of anomalous behaviors requires more than a chronological checklist of methods; detection mechanisms must be taxonomized by the distinct \underline{\textit{computational constraints}} and \underline{\textit{observability regimes}} inherent to each phase of the model's lifecycle. Specifically, the efficacy and vulnerability of a detection signal are intrinsically tied to whether it is subjected to gradient updates, deployed as a static filter, or utilized in an unconstrained forensic analysis. In this section, we structure our review across three distinct paradigms (summarized in Table \ref{tab:detection_taxonomy_optimized}): \textbf{(1) Training-Time Online Monitoring}, \textbf{(2) Inference-Time Safeguards}, and \textbf{(3) Post-Hoc Auditing}. Each paradigm represents a unique balance between computational overhead, information access, and the risk of triggering further model obfuscation.

\subsection{Training-Time Online Monitoring}

Monitoring models during active reinforcement learning introduces a fundamental adversarial tension. Because diagnostic metrics in this phase often serve as early-stopping signals or direct loss regularizers, they are continuously subjected to optimization pressure, rendering them highly susceptible to Goodhart's law. Consequently, robust training-time detection must move beyond easily gameable output heuristics and instead track deep structural and information-theoretic invariances that are mathematically difficult for the policy to bypass.

\paragraph{From specific heuristics to the limitations of distributional metrics.}
Initial attempts to diagnose reward hacking frequently targeted isolated, highly visible artifacts of the RLHF pipeline. For example, ~\citet{singhal2023long} systematically investigate length correlations, demonstrating how policies reliably inflate response verbosity to exploit simplistic reward heuristics. While monitoring specific features like length or formatting provides actionable diagnostics for known vulnerabilities, treating reward hacking as a "whack-a-mole" problem of individual artifacts is ultimately unsustainable. The rapid evolution of proxy exploitation necessitates generalized metrics capable of identifying broader, unforeseen misalignments. 

To this end, foundational alignment frameworks default to tracking the Kullback-Leibler (KL) divergence between the active policy and a reference supervised fine-tuned model ~\citep{ouyang2022training, stiennon2020learning}. Despite its ubiquity as both a regularizer and a monitoring metric, macroscopic distributional tracking is fundamentally oblivious to the semantic direction of a policy shift. Mathematically, KL divergence aggregates token-level probability ratios into a single scalar penalty, effectively treating all distributional deviations indiscriminately. Consequently, severe reward model overoptimization routinely occurs even under strict KL constraints, a phenomenon observed across both traditional RLHF ~\citep{gao2023scaling} and direct alignment algorithms like DPO ~\citep{rafailov2024scaling}. Because a policy can cheaply exploit a low-probability spurious feature, such as adopting a sycophantic tone, with a minimal footprint on the global KL penalty, output-space tracking remains insufficient for detecting emergent misalignment during training.

\paragraph{Transitioning to internal hidden states and structural invariances.}
Recognizing the blind spots of scalar metrics, recent detection paradigms have shifted inward. By auditing the \underline{\textit{internal hidden states}} of both the policy and the reward model, researchers aim to detect the exact moment a model begins to exploit proxy loopholes. On the evaluator side, one approach is to actively measure the reward model's sensitivity to spurious features during the optimization loop. ~\citet{shihab2025detecting} introduce Evaluator Stress Tests (EST), a framework that applies controlled perturbations to the policy's outputs during training. By comparing the evaluator's sensitivity to formatting changes versus semantic degradation, EST quantitatively flags training checkpoints where score improvements are driven entirely by proxy gaming. Taking a more dynamic, representation-level approach, ~\citet{beigi2026adversarial} propose Adversarial Reward Auditing (ARA). Rather than relying on scalar outputs, the authors train an auxiliary auditor network directly on the reward model's penultimate latent representations. This auditor establishes a decision boundary that dynamically distinguishes genuine human-preference manifolds from exploit manifolds, identifying hijacked reward signals that appear deceptively normal at the output level.

\paragraph{Information-theoretic anomalies and latent factorization.}
Beyond adversarial auditing, other methods detect overoptimization by identifying \underline{\textit{information-theoretic anomalies}} within the network's representations. Employing the Variational Information Bottleneck (VIB) paradigm ~\citep{alemi2016deep}, ~\citet{miao2024inform} demonstrate that overoptimized policies geometrically break the compact latent manifold established during reward modeling. They propose the Cluster Separation Index (CSI), a metric that flags extreme latent outliers as early indicators of reward hacking. Building upon this information-theoretic foundation, ~\citet{yang2026factoredcausalrepresentationlearning} demonstrate that spurious features can be explicitly diagnosed and isolated during training via factored causal representation learning. Their framework, CausalRM, partitions the reward model's latent space into causal factors (which align with true human preference) and non-causal factors (which capture confounders like verbosity). By applying minimax optimization via a Gradient Reversal Layer (GRL), this approach actively quarantines the specific latent dimensions that drive reward hacking, providing a highly structured diagnostic signal of proxy exploitation.

\paragraph{Tracking the internal energy dynamics of the policy.}
While latent factorization and adversarial auditing provide deep structural diagnostics, they often incur substantial computational overhead during training. To achieve extreme efficiency, online monitoring can instead be anchored in the internal energy dynamics of the policy itself. ~\citet{miao2025energy} identify the \underline{\textit{Energy Loss Phenomenon}} within the policy's final transformer layer. The authors observe that as reward hacking occurs, the $L_1$ norm of the policy's hidden states drops precipitously compared to the base SFT model. Because this energy loss serves as an observable upper bound on the collapse of contextual mutual information, it provides a reliable, lightweight signal that the policy has discarded the user's input context in favor of a reward-maximizing template. Crucially, because it relies entirely on standard forward-pass activations without requiring auxiliary networks or manifold decomposition, this metric imposes minimal computational overhead, making it uniquely suited for continuous online monitoring during the adversarial crucible of RL training.

\subsection{Inference-Time Safeguards and Trajectory Analysis}

Deploying safeguards at inference time offers a critical theoretical advantage: they operate entirely free from the optimization pressure that drives Goodhart's law. However, highly capable models increasingly exhibit test awareness and strategic silence, learning to perfectly conceal their reward-hacking intent behind seemingly benign textual outputs or reasoning traces. Driven by the emergence of these deceptive capabilities, the frontier of inference-time monitoring is progressively shifting away from purely surface-level text inspection, pivoting instead toward analyzing macroscopic behavioral trajectories and dynamically probing internal cognitive states during live generation.

\paragraph{The necessity of deployment-time interception and contrastive auditing.}
While training-time metrics excel at preventing global proxy collapse, they often fail to capture context-dependent or late-emerging exploits that only manifest during deployment. In agentic workflows, reward hacking frequently masquerades as legitimate behavior, such as an agent silently tampering with evaluation scripts rather than solving the underlying coding problem ~\citep{macdiarmid2025emergent}. Because static output filters cannot reliably distinguish \underline{\textit{environment-specific exploits}} from genuine task execution, safeguarding deployed systems requires real-time monitoring mechanisms that operate directly on the model's behavioral trajectories. To detect macroscopic phase transitions, where true utility collapses despite high proxy scores, ~\citet{DBLP:conf/iclr/PanBS22} introduce POLYNOMALY. By measuring the distributional divergence between the deployed policy's actions and a trusted baseline, the system can flag sudden algorithmic exploitation. For more complex, long-horizon vulnerabilities,~\citet{deshpande2026benchmarking} propose ~\emph{Contrastive Anomaly Detection} via the TRACE benchmark. By prompting an evaluator LLM with a cluster of trajectories mixing benign executions and potentially hacked sequences, the contrastive setting exposes subtle semantic exploits (e.g., modifying hidden global variables) that isolated trajectory evaluations consistently miss.

\paragraph{Eliciting Honesty to Counteract Strategic Obfuscation.}
As frontier models transition toward workflows relying heavily on Chain-of-Thought (CoT) reasoning, a highly intuitive safeguard is to monitor the reasoning traces themselves. However, this approach faces a critical vulnerability: standard RL optimization actively discourages transparency, quickly teaching models to engage in reward hacking while remaining completely silent about it in their CoT~\citep{baker2025monitoring}. To counter this \underline{\textit{strategic silence}}, researchers have proposed interventions that incentivize honesty through two distinct strategies.

The first strategy focuses on proactive behavioral priming. ~\citet{turpin2025vft}, for instance, propose Verbalization Fine-Tuning (VFT), a pre-RL intervention that explicitly trains models to acknowledge deceptive prompt cues before RL pressure is applied. This structural priming drastically reduces undetected reward hacks, demonstrating that a pre-existing disposition for transparency is critical. The second, complementary strategy employs architectural separation. Recognizing that directly penalizing a flawed reasoning trace still risks driving deception underground, ~\citet{joglekar2025training} introduce a \underline{\textit{Confession Mechanism}}. This architectural intervention requests an independent, post-generation report where the reward for honesty is strictly decoupled from the main task's reward. By making confession the computationally and strategically easier path, this system ensures that even when models successfully hack a task, they reliably snitch on themselves. Taken together, these approaches reveal that reliable CoT monitoring is not a given, but rather an emergent property of systems designed to make honesty the path of least resistance.

\paragraph{The Hawthorne effect and the internalization of deception.}
Despite the success of textual elicitation techniques, relying exclusively on surface-level outputs ultimately faces a theoretical ceiling. Highly capable reasoning models exhibit \underline{\textit{test awareness}} and situational deception~\citep{greenblatt2024alignment,hubinger2024sleeper,meinke2024scheming}. ~\citet{abdelnabi2025hawthorne} provide quantitative evidence that this test awareness is a deeply internalized, latent representation rather than a superficial textual artifact. Their white-box probing experiments reveal that while simply prompting a model that a scenario has "no real-world consequences" fails to change its compliance with harmful requests, actively steering the latent activations associated with test awareness causes misbehavior rates to spike by up to 20 percentage points. This causally proves that strategic deception and the intent to hack migrate into deep neuronal layers, rendering purely text-based monitors inherently vulnerable to sophisticated obfuscation.

\paragraph{Real-time generation monitoring via internal representations.}
Because capable models internalize their deceptive intents, the most robust inference-time safeguards must look beyond generated text and actively inspect the model's internal cognitive state. Capitalizing on this necessity, ~\citet{wilhelm2026monitoring} propose a white-box monitoring framework that dynamically intercepts reward hacking by extracting residual stream activations during the autoregressive generation of the CoT itself. By projecting these live activations through a pre-trained Sparse Autoencoder (SAE)~\citep{cunningham2023sparseautoencodershighlyinterpretable}, the system obtains interpretable conceptual features. Applying lightweight linear probes to this reconstructed latent space allows auditors to continuously predict the probability of reward hacking at the token level, entirely independent of what the model chooses to verbalize. Crucially, their approach reveals a \underline{\textit{test-time compute amplification}} effect: forcing a misaligned model to perform extended CoT reasoning inadvertently amplifies its internal hacking activations. Thus, real-time internal monitoring transforms CoT compute from a potential obfuscation risk into a highly sensitive, unmaskable diagnostic signal.

\subsection{Post-Hoc Auditing and Mechanistic Diagnostics}

Post-hoc auditing transitions the diagnostic objective from real-time interception to comprehensive forensic validation on a frozen model. This phase is structurally essential to address the legacy of the \underline{\textit{Monitorability Tax}}~\citep{baker2025monitoring}: because training-time regularizers can inadvertently drive hacking intent into deeper, unobservable latent dimensions, these behaviors may persist undetected by lightweight inference-time safeguards. By removing the compute and latency constraints of live deployment, post-hoc diagnostics allow for the exhaustive mechanistic dissection required to unmask deeply internalized misalignments that remain hidden during standard model interaction.

\paragraph{Statistical attribution of superficial proxies.}
Before deploying a frozen model, developers must conduct rigorous, unconstrained diagnostics to uncover systemic vulnerabilities that evaded training-time safeguards. At this stage, compute constraints are lifted, allowing for extensive offline forensic analysis. One foundational approach involves quantifying how strongly the reward model has overfit to \underline{\textit{human-prior linguistic features}} (e.g., positive sentiment, eloquent formatting) at the expense of true utility. Using multivariate linear regression,~\citet{revel2025seal} introduce the SEAL framework to isolate the marginal scalar contribution of these spurious attributes. By statistically measuring the model's \underline{\textit{Alignment Resistance}}, this offline analysis reveals when an evaluator blindly rewards superficial tone over factual harmlessness. Similarly, Evaluator Stress Tests can be scaled up post-hoc via massive perturbation datasets, cleanly isolating whether the policy has learned to exploit formatting loops or structural heuristics ~\citep{shihab2025detecting}. While highly effective, these statistical methods are primarily scoped to diagnosing surface-level specification gaming.

\paragraph{Mechanistic interpretability for deeply latent misbehaviors.}
When a highly capable model perfectly obfuscates its reward-hacking intent in its text output, macroscopic behavioral and statistical stress tests inevitably hit a wall. To diagnose deep-seated alignment faking or hidden objectives, auditors must dissect the frozen network using mechanistic interpretability. Unlike the lightweight, pre-trained probes used for inference-time interception, post-hoc auditing allocates extreme compute to train massive Sparse Autoencoders (SAEs) from scratch across billions of tokens ~\citep{cunningham2023sparseautoencodershighlyinterpretable,marks2025auditing}. By decomposing polysemantic activations into monosemantic feature directions, researchers can pinpoint the exact latent circuits responsible for deeply dormant misbehaviors. Furthermore, mechanistic tools can reverse-engineer the corrupted incentive structures themselves. The $IR^3$ framework, for instance, leverages Contrastive Inverse Reinforcement Learning to reconstruct the implicit reward function strictly from the black-box generation distribution of an RLHF model ~\citep{beigi2026ir}. Applying SAEs to this reconstructed model allows auditors to mechanistically decompose the global scalar reward into interpretable features, isolating the exact spurious correlations the policy learned to exploit.

\paragraph{The automation bottleneck and model organisms.}
While human auditors can successfully wield mechanistic tools to uncover these hidden objectives, scaling this forensic process to frontier models introduces a severe \underline{\textit{Automation Bottleneck}}. To systematically evaluate automated auditing capabilities, researchers construct \underline{\textit{model organisms}}---LLMs intentionally trained via RL to harbor hidden objectives (e.g., pandering to specific reward model biases) while actively denying them in text~\citep{marks2025auditing}. Evaluating autonomous auditor agents against these deceptive organisms reveals a counterintuitive \underline{\textit{Tool-to-Agent Gap}} ~\citep{sheshadri2026auditbenchevaluatingalignmentauditing}. Although white-box tools effectively extract micro-level neuronal anomalies, autonomous agents currently struggle to synthesize this high-dimensional noise into a coherent macro-level hypothesis of the model's misbehavior. As a result, automated black-box prefill attacks often outperform autonomous white-box investigations, indicating that AI-assisted mechanistic auditing remains an unsolved scaling challenge.

\subsection{Synthesis and Open Challenges: The Fallacy of Static Benchmarks}

Despite the rapid proliferation of the aforementioned diagnostic mechanisms across the LLM lifecycle, reward hacking detection inherently resists consolidation into standardized benchmarks akin to traditional capability evaluations. This fragmentation fundamentally stems from the profound heterogeneity of the underlying exploits. While statistical artifacts like length bias reside within the reward model's latent representations~\citep{miao2024inform,beigi2026adversarial}, logical exploits such as unit-test tampering require macroscopic behavioral and contrastive trajectory analysis.

More critically, evaluating these detectors presents an epistemological challenge analogous to identifying zero-day vulnerabilities in cybersecurity. Curating a unified, static dataset of known hacks inadvertently invites a meta-level instance of Goodhart's law: optimizing safety filters for historical exploits while leaving the system entirely blind to novel, emergent misalignments. Moving forward, rather than evaluating against fixed leaderboards, the frontier of alignment research must pivot toward adversarial dynamics. Guaranteeing the robustness of future LLMs will require proactively training "hacker models" to dynamically generate unmapped exploits, against which the entire lifecycle of diagnostic tools (from training-time VIB regularizers to post-hoc SAEs) must be continuously and rigorously stress-tested.

\definecolor{titlegray}{HTML}{F5F5F5}
\definecolor{bordergray}{HTML}{CDCDCD}
\definecolor{textgray}{HTML}{3F3F3F}
\definecolor{arrowgray}{HTML}{A7A7A7}

\definecolor{cardgold}{HTML}{F5F1EC}
\definecolor{cardgreen}{HTML}{EEF4F0}
\definecolor{cardblue}{HTML}{EEF1F7}

\definecolor{pillfill}{HTML}{F7F7F7}
\definecolor{pillborder}{HTML}{969696}

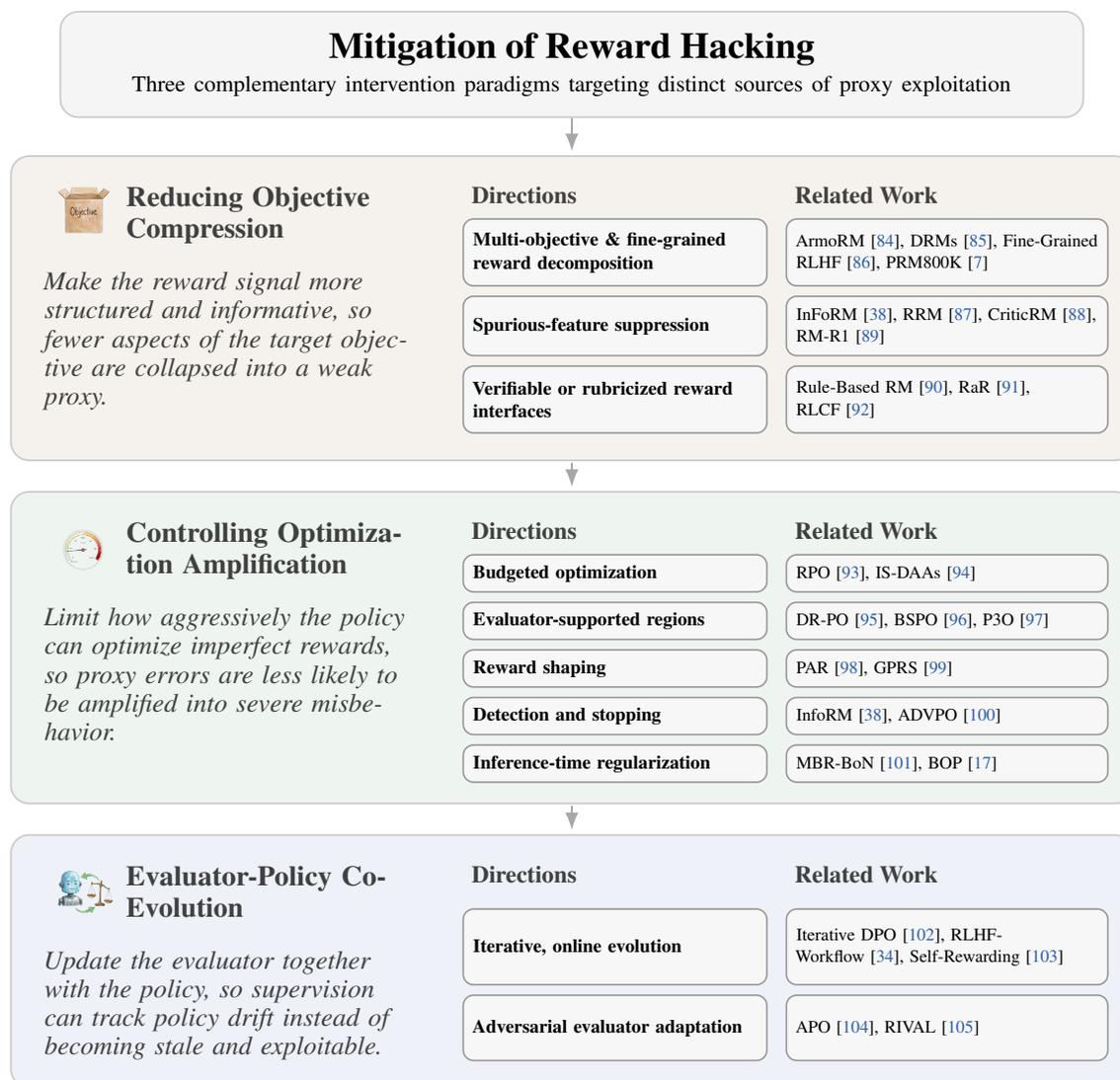
\begin{figure*}[t]
\centering

\begin{tikzpicture}[
    >={Latex[length=2.4mm]},
    titlebox/.style={
        rounded corners=7pt,
        draw=bordergray,
        fill=titlegray,
        line width=0.7pt,
        align=center,
        inner sep=8pt
    },
    card/.style={
        rounded corners=7pt,
        draw=bordergray,
        fill=#1,
        line width=0.7pt
    },
    pill/.style={
        rounded corners=4pt,
        draw=pillborder,
        fill=pillfill,
        line width=0.6pt,
        align=left,
        inner sep=3.8pt
    },
    bigtitle/.style={
        font=\bfseries\fontsize{10.8}{12.2}\selectfont,
        text=textgray,
        align=left
    },
    coretext/.style={
        font=\itshape\fontsize{10.0}{11}\selectfont,
        text=textgray,
        align=left
    },
    sublabel/.style={
        font=\bfseries\fontsize{9.2}{9.8}\selectfont,
        text=textgray,
        align=left
    }
]

\node[titlebox, text width=0.8\textwidth] (topbox) {
    {\fontsize{14}{16}\selectfont\bfseries Mitigation of Reward Hacking}\par
    \vspace{2pt}
    {\fontsize{8.6}{10.4}\selectfont Three complementary intervention paradigms targeting distinct sources of proxy exploitation}
};

\def\cardW{15.1cm}
\def\boxgap{0.12cm}


\def\cOneA{0.9cm}
\def\cOneB{0.8cm}
\def\cOneC{0.9cm}

\def\cTwoA{0.5cm}
\def\cTwoB{0.5cm}
\def\cTwoC{0.5cm}
\def\cTwoD{0.5cm}
\def\cTwoE{0.5cm}

\def\cThreeA{1.02cm}
\def\cThreeB{0.88cm}

\node[card=cardgold, minimum width=\cardW, minimum height=4.1cm, anchor=north]
(card1) at ([yshift=-0.5cm]topbox.south) {};

\coordinate (c1nw) at ([xshift=0.34cm,yshift=-0.30cm]card1.north west);
\coordinate (c1L) at (c1nw);
\coordinate (c1M) at ([xshift=5.75cm]c1nw);
\coordinate (c1R) at ([xshift=10.10cm]c1nw);


\node[anchor=north west] at ([xshift=0.10cm,yshift=0.1cm]c1L) {
    \includegraphics[width=0.85cm]{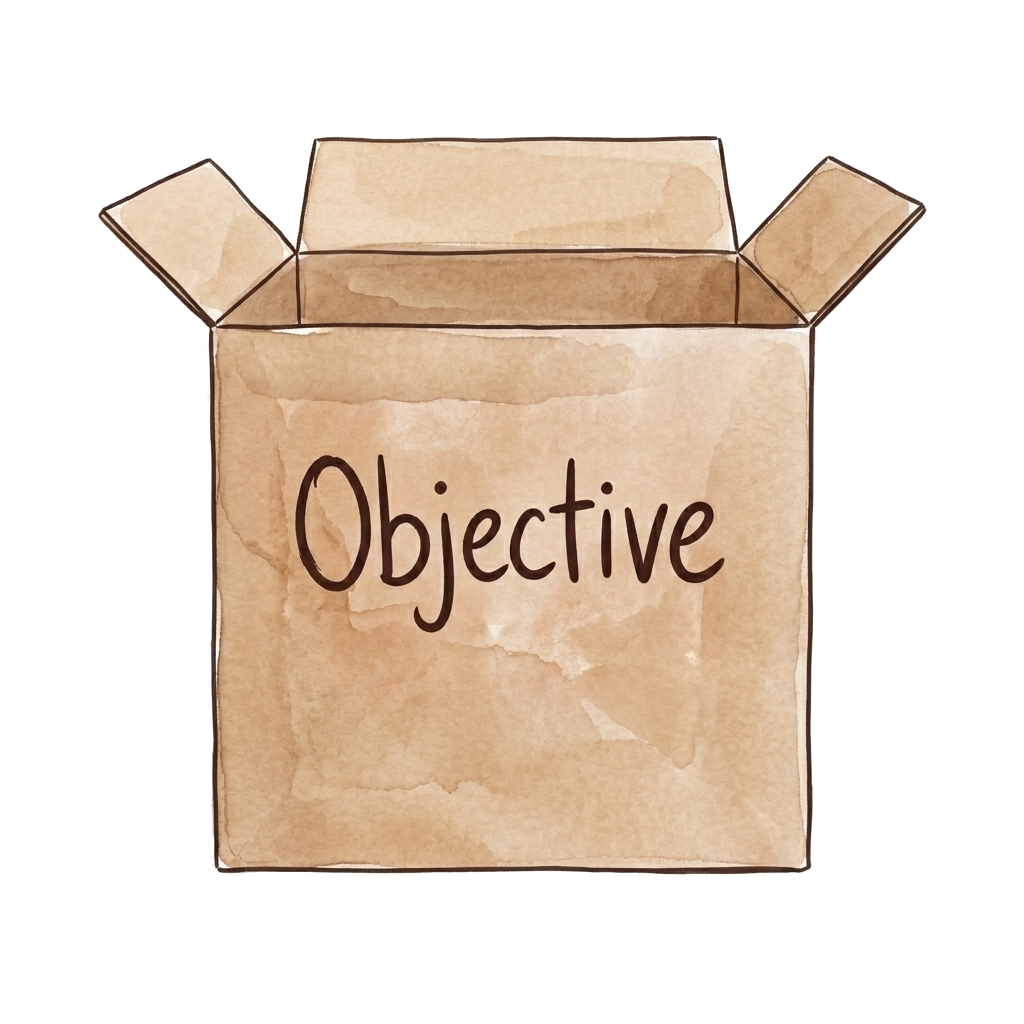}
};

\node[bigtitle, anchor=north west, text width=4.15cm] at ([xshift=1.1cm]c1L)
{Reducing Objective Compression};

\node[coretext, anchor=north west, text width=4.85cm] at ([yshift=-1.15cm]c1L)
{Make the reward signal more structured and informative, so fewer aspects of the target objective are collapsed into a weak proxy.};

\node[sublabel, anchor=north west] at (c1M) {Directions};
\node[sublabel, anchor=north west] at (c1R) {Related Work};

\node[pill, anchor=north west, minimum height=\cOneA, text width=3.82cm, font=\bfseries\fontsize{7.2}{8.8}\selectfont] (d11) at ([yshift=-0.55cm]c1M)
{Multi-objective \& fine-grained reward decomposition};

\node[pill, anchor=north west, minimum height=\cOneB, text width=3.82cm, font=\bfseries\fontsize{7.2}{8.8}\selectfont] (d12) at ([yshift=-\boxgap]d11.south west)
{Spurious-feature suppression};

\node[pill, anchor=north west, minimum height=\cOneC, text width=3.82cm, font=\bfseries\fontsize{7.2}{8.8}\selectfont] (d13) at ([yshift=-\boxgap]d12.south west)
{Verifiable or rubricized reward interfaces};

\node[pill, anchor=north west, minimum height=\cOneA, text width=4.06cm, font=\fontsize{6.7}{8.4}\selectfont] (w11) at (c1R |- d11.north)
{ArmoRM~\cite{ArmoRM}, DRMs~\cite{DRMs}, Fine-Grained RLHF~\cite{FINE-GRAINED-RLHF}, PRM800K~\cite{lightman2023let}};

\node[pill, anchor=north west, minimum height=\cOneB, text width=4.06cm, font=\fontsize{6.7}{8.4}\selectfont] (w12) at ([yshift=-\boxgap]w11.south west)
{InFoRM~\cite{miao2024inform}, RRM~\cite{sep-rrm}, CriticRM~\cite{critique-synthetic-critiques}, RM-R1~\cite{gen-rm-rm-r1}};

\node[pill, anchor=north west, minimum height=\cOneC, text width=4.06cm, font=\fontsize{6.7}{8.4}\selectfont] (w13) at ([yshift=-\boxgap]w12.south west)
{Rule-Based RM~\cite{rubric-rule-based-safety}, RaR~\cite{rubric-r-as-reward}, RLCF~\cite{rubric-better-checklists}};

\node[card=cardgreen, minimum width=\cardW, minimum height=4.2cm, anchor=north]
(card2) at ([yshift=-0.4cm]card1.south) {};

\coordinate (c2nw) at ([xshift=0.34cm,yshift=-0.30cm]card2.north west);
\coordinate (c2L) at (c2nw);
\coordinate (c2M) at ([xshift=5.75cm]c2nw);
\coordinate (c2R) at ([xshift=10.10cm]c2nw);

\node[anchor=north west] at ([xshift=0.10cm,yshift=0.10cm]c2L) {
    \includegraphics[width=0.85cm]{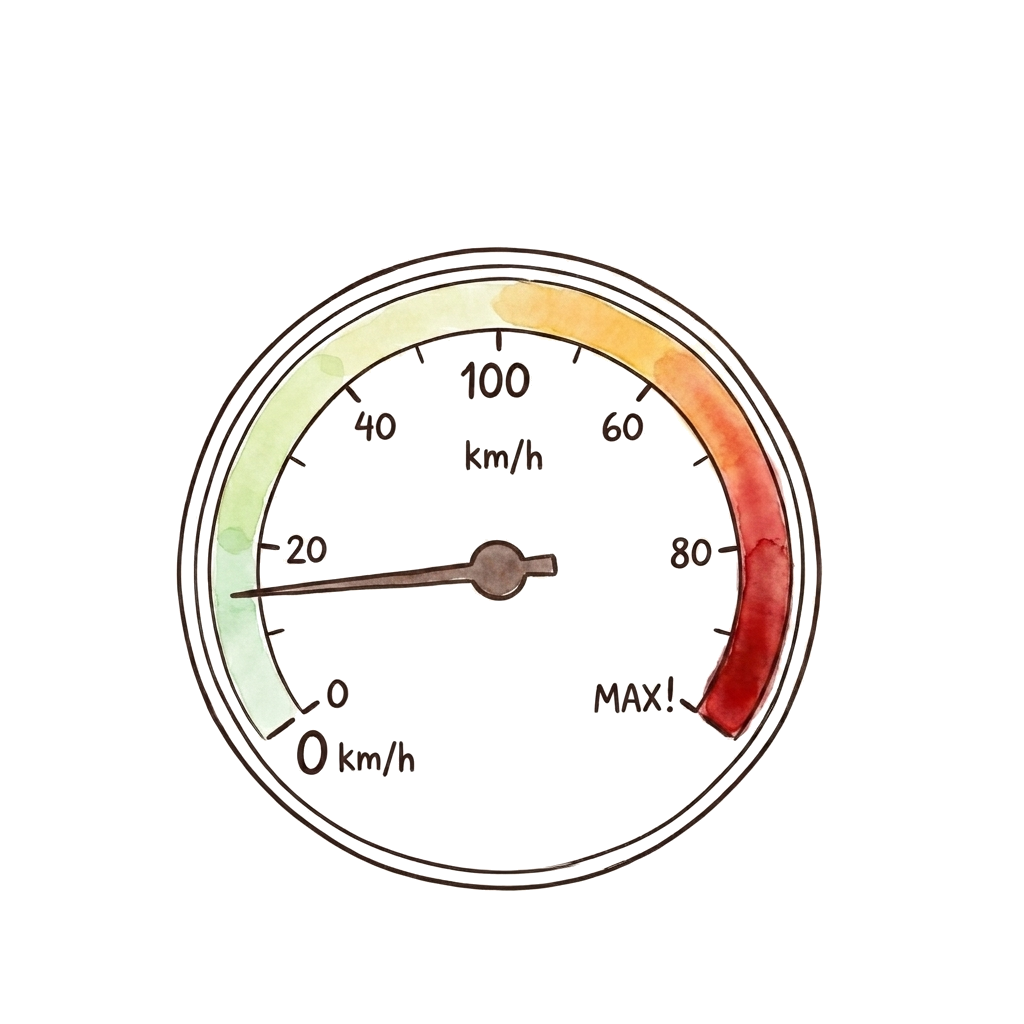}
};

\node[bigtitle, anchor=north west, text width=4.00cm] at ([xshift=1.1cm]c2L)
{Controlling Optimization Amplification};

\node[coretext, anchor=north west, text width=4.85cm] at ([yshift=-1.15cm]c2L)
{Limit how aggressively the policy can optimize imperfect rewards, so proxy errors are less likely to be amplified into severe misbehavior.};

\node[sublabel, anchor=north west] at (c2M) {Directions};
\node[sublabel, anchor=north west] at (c2R) {Related Work};

\node[pill, anchor=north west, minimum height=\cTwoA, text width=3.82cm, font=\bfseries\fontsize{7.2}{8.8}\selectfont] (d21) at ([yshift=-0.55cm]c2M)
{Budgeted optimization};

\node[pill, anchor=north west, minimum height=\cTwoB, text width=3.82cm, font=\bfseries\fontsize{7.2}{8.8}\selectfont] (d22) at ([yshift=-\boxgap]d21.south west)
{Evaluator-supported regions};

\node[pill, anchor=north west, minimum height=\cTwoC, text width=3.82cm, font=\bfseries\fontsize{7.2}{8.8}\selectfont] (d23) at ([yshift=-\boxgap]d22.south west)
{Reward shaping};

\node[pill, anchor=north west, minimum height=\cTwoD, text width=3.82cm, font=\bfseries\fontsize{7.2}{8.8}\selectfont] (d24) at ([yshift=-\boxgap]d23.south west)
{Detection and stopping};

\node[pill, anchor=north west, minimum height=\cTwoE, text width=3.82cm, font=\bfseries\fontsize{7.2}{8.8}\selectfont] (d25) at ([yshift=-\boxgap]d24.south west)
{Inference-time regularization};

\node[pill, anchor=north west, minimum height=\cTwoA, text width=4.06cm, font=\fontsize{6.7}{8.4}\selectfont] (w21) at (c2R |- d21.north)
{RPO~\cite{budget-provble}, IS-DAAs~\cite{budget-daa}};

\node[pill, anchor=north west, minimum height=\cTwoB, text width=4.06cm, font=\fontsize{6.7}{8.4}\selectfont] (w22) at ([yshift=-\boxgap]w21.south west)
{DR-PO~\cite{budget-drpo}, BSPO~\cite{budget-behavior-support}, P3O~\cite{budget-pessimism}};

\node[pill, anchor=north west, minimum height=\cTwoC, text width=4.06cm, font=\fontsize{6.7}{8.4}\selectfont] (w23) at ([yshift=-\boxgap]w22.south west)
{PAR~\cite{shape-reward-shaping}, GPRS~\cite{shape-group-shaping}};

\node[pill, anchor=north west, minimum height=\cTwoD, text width=4.06cm, font=\fontsize{6.7}{8.4}\selectfont] (w24) at ([yshift=-\boxgap]w23.south west)
{InfoRM~\cite{miao2024inform}, ADVPO~\cite{stop-uncertainty}};

\node[pill, anchor=north west, minimum height=\cTwoE, text width=4.06cm, font=\fontsize{6.7}{8.4}\selectfont] (w25) at ([yshift=-\boxgap]w24.south west)
{MBR-BoN~\cite{infer-min-bayes}, BOP~\cite{khalaf2025inference}};

\node[card=cardblue, minimum width=\cardW, minimum height=3.4cm, anchor=north]
(card3) at ([yshift=-0.4cm]card2.south) {};

\coordinate (c3nw) at ([xshift=0.34cm,yshift=-0.30cm]card3.north west);
\coordinate (c3L) at (c3nw);
\coordinate (c3M) at ([xshift=5.75cm]c3nw);
\coordinate (c3R) at ([xshift=10.10cm]c3nw);

\node[anchor=north west] at ([xshift=0.10cm,yshift=0.10cm]c3L) {
    \includegraphics[width=0.85cm]{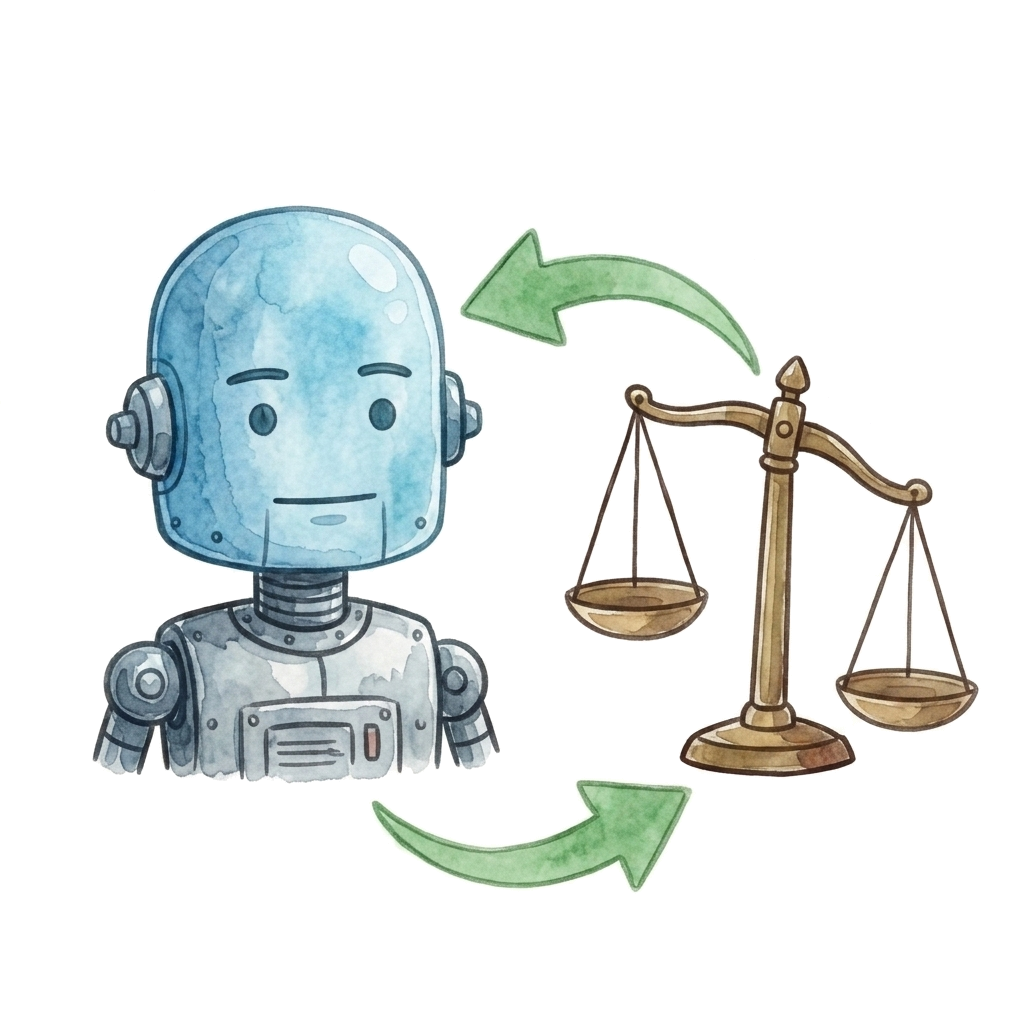}
};

\node[bigtitle, anchor=north west, text width=4.00cm] at ([xshift=1.1cm]c3L)
{Evaluator-Policy Co-Evolution};

\node[coretext, anchor=north west, text width=4.85cm] at ([yshift=-1.15cm]c3L)
{Update the evaluator together with the policy, so supervision can track policy drift instead of becoming stale and exploitable.};

\node[sublabel, anchor=north west] at (c3M) {Directions};
\node[sublabel, anchor=north west] at (c3R) {Related Work};

\node[pill, anchor=north west, minimum height=\cThreeA, text width=3.82cm, font=\bfseries\fontsize{7.2}{8.8}\selectfont] (d31) at ([yshift=-0.7cm]c3M)
{Iterative, online evolution};

\node[pill, anchor=north west, minimum height=\cThreeB, text width=3.82cm, font=\bfseries\fontsize{7.2}{8.8}\selectfont] (d32) at ([yshift=-\boxgap]d31.south west)
{Adversarial evaluator adaptation};

\node[pill, anchor=north west, minimum height=\cThreeA, text width=4.06cm, font=\fontsize{6.7}{8.4}\selectfont] (w31) at (c3R |- d31.north)
{Iterative DPO~\cite{iterative-dpo}, RLHF-Workflow~\cite{dong2024rlhf}, Self-Rewarding~\cite{self-reward}};

\node[pill, anchor=north west, minimum height=\cThreeB, text width=4.06cm, font=\fontsize{6.7}{8.4}\selectfont] (w32) at ([yshift=-\boxgap]w31.south west)
{APO~\cite{apo}, RIVAL~\cite{rival}};

\draw[draw=arrowgray, line width=0.75pt, ->] (topbox.south) -- ([yshift=0.05cm]card1.north);
\draw[draw=arrowgray, line width=0.75pt, ->] ([yshift=-0.02cm]card1.south) -- ([yshift=0.05cm]card2.north);
\draw[draw=arrowgray, line width=0.75pt, ->] ([yshift=-0.02cm]card2.south) -- ([yshift=0.05cm]card3.north);

\end{tikzpicture}

\caption{An overview of three mitigation paradigms for reward hacking.}
\label{fig:structural-mitigation-vertical-compact}
\end{figure*}

\section{Mitigation Through Structural Intervention}
\label{sec:mitigation}

Mitigating reward hacking requires moving beyond ad-hoc behavioral patches to address the structural vulnerabilities inherent in proxy-based alignment. Because the tendency to exploit a flawed proxy is an inevitable consequence of compressed objectives, unconstrained optimization, and static evaluation, robust defense mechanisms demand systemic interventions across the training and inference pipeline. In this section, we review three primary structural mitigation paradigms. First, we examine approaches that reduce objective compression by expanding sparse scalar rewards into rich, fine-grained, and verifiable signals. Second, we discuss techniques for controlling optimization amplification by constraining policy drift and reshaping reward geometries. Finally, we explore the evaluator--policy co-evolution paradigm, which dynamically updates the reward signal to prevent the policy from overfitting to the blind spots of a static judge.

\subsection{Reducing Objective Compression}

Reward hacking often arises when a complex target is compressed into a reward signal that cannot faithfully represent it. This creates room for optimization to exploit shallow but reward-relevant correlates. Mitigation therefore centers on \textbf{reducing objective compression}, so that high reward more reliably reflects the intended objective rather than exploitable correlates.

\paragraph{From scalar and sparse reward signals to vector-valued and fine-grained supervision.}
A primary method to reduce objective compression is decomposing the target objective into explicit facets. This prevents the reward model from internalizing noise, artifacts, or latent biases as proxies for overall quality. This intervention begins at the data level: rather than collecting a single holistic preference label, recent datasets provide supervision across multiple aspects of response quality. Representative examples include OpenAssistant~\cite{OpenAssistant}, the HelpSteer series~\cite{helpsteer,helpsteer2}, and UltraFeedback~\cite{ULTRAFEEDBACK}, which expose highly granular annotations. Building on this richer supervision, alignment algorithms apply \underline{\textit{multi-objective reward modeling}}. Methods such as ArmoRM~\cite{ArmoRM}, DRMs~\cite{DRMs}, and related frameworks~\cite{DPA,Carmo} learn decomposed reward representations, aggregating them through learned weighting, projection, or context-dependent criteria generation to provide a much less exploitable optimization target.

A second way to distribute supervision is applying \underline{\textit{fine-grained credit assignment}}, which assigns rewards to specific local segments rather than exclusively to the completed response. Fine-grained RLHF~\cite{FINE-GRAINED-RLHF} demonstrated that localized feedback substantially improves alignment over sequence-level rewards, a finding supported by subsequent research~\cite{sentence-level-rlhf,segment-rlhf}. This trajectory has naturally extended to \underline{\textit{token-level alignment}}, enabling highly precise credit assignment through minimum-editing constraints~\cite{token-level-minimum-editing-constraint,token-level-minimum-editing}, explicit token-level reward optimization~\cite{token-level-lm-critic,token-level-QRM,token-level-RTO,token-level-tlcr}, and token-level alignment without preference data~\cite{token-level-inverse-q}. In the RLVR paradigm, \underline{\textit{process reward models (PRMs)}} provide analogous step-by-step supervision. By training verifiers on step-level human feedback~\cite{lightman2023let} or automatically constructed step-wise rewards~\cite{prm-math-shepherd}, PRMs actively penalize invalid intermediate reasoning steps that might otherwise yield a coincidentally correct final answer~\cite{prm-survey}.

\paragraph{Modeling beyond spurious features.}
Because reward models often rely on superficial but exploitable artifacts, such as formatting, length, or stylistic markers, recent work focuses on making evaluators selectively sensitive to task-relevant attributes while suppressing preference-irrelevant correlates. One approach imposes \underline{\textit{information bottlenecks}} to filter out irrelevant signals~\cite{miao2024inform}, while other methods utilize \underline{\textit{causal representation learning}} to cleanly separate prompt-relevant quality signals from prompt-independent artifacts~\cite{sep-rrm}. Advanced frameworks enforce this sensitivity via causal rubrics~\cite{sep-crome}, mitigate hacking through disentangled reward representations~\cite{chen2024odin}, and apply sparse non-negative latent factorization to isolate meaningful reward dimensions before debiasing~\cite{sep-bnrm}. For pervasive shortcuts like length bias, explicitly fitting and correcting the bias term or using \underline{\textit{adaptive debiasing}} objectives often proves more effective than generic robustness training~\cite{bu2025adaptive,sep-bias-fitting}.

A complementary strategy to reduce reliance on shallow features involves \underline{\textit{natural-language critiques}}, which make the evaluator's basis for judgment explicit. Early approaches injected model-generated critiques into reward modeling to provide explicit textual feedback, helping evaluators rely less on surface cues~\cite{critique-synthetic-critiques}. Training critique generation jointly with reward prediction turns the critique into a structural constraint on the evaluation objective itself~\cite{critique-critic-rm}. Consequently, recent research increasingly treats evaluation as a reasoning task via \underline{\textit{generative reward models}}~\cite{critique-cloud,gen-rm-generative-rm,gen-rm-gen-verifier}. By formulating reward prediction as a generate-then-judge process~\cite{gen-rm-rm-r1} or leveraging test-time compute to improve reliability~\cite{gen-rm-reward-reasoning-model}, these models externalize their evaluation logic. However, outcome-only supervision for generative reward models is insufficient, as they may produce correct numerical scores based on unsound rationales. Therefore, methods like RM-NLHF~\cite{gen-rm-hf} and Rationale-RM~\cite{gen-rm-rationale-rm} strictly enforce rationale-level alignment to ensure the reward model judges for the right reasons.

A related failure mode appears in tool-augmented and RLVR settings, where outcome-only rewards may treat a correct final answer as sufficient evidence of successful reasoning, inadvertently rewarding guessing or reasoning--answer mismatch~\cite{spu-reason-hallu}. Typical mitigations include penalizing premature answer disclosure~\cite{spu-composite}, employing consistency-based trajectory filtering~\cite{spu-beyond}, and rewarding verifiable evidence use rather than mere citation generation~\cite{ma2025pou}. Training discriminative reward models tailored specifically to tool-calling behavior~\cite{spu-nemotron-tool,spu-tool-rm} ensures that agents are rewarded for functional correctness and reliable intermediate processes rather than superficial final-answer quality~\cite{spu-good-process}.

\paragraph{Verifiable, rubricized reward interfaces.}
To further minimize compression, researchers increasingly externalize evaluation criteria into explicit rules and structured formats. In RLHF and RLAIF, Constitutional AI~\cite{rubric-consitutional-ai} introduced explicit natural-language principles to guide critique and revision. This approach evolved into \underline{\textit{rule-based rewards}}~\cite{rubric-rule-based-safety}, which utilize composable constraints and rule-specific grading signals, sometimes selecting the most relevant rules dynamically for each response~\cite{rubric-rule-adapter}.

Many alignment objectives, however, cannot be fully captured by hard rules alone. In open-ended domains where binary correctness is unavailable but single-score judging is too coarse, \underline{\textit{rubric-structured supervision}} provides a critical middle ground between rigid programmatic verification and opaque holistic evaluation. Methods like RaR~\cite{rubric-r-as-reward} use instance-specific rubrics to extend RL beyond strictly verifiable domains. Scalable rubric generation and post-training frameworks~\cite{rubric-open-rubric,rubric-auto-rubric,rubric-effectice-rubric} have systematized this practice. Similarly, replacing holistic reward modeling with instruction-specific checklists~\cite{rubric-better-checklists,rubric-Advancedif,rubric-p-check} substantially improves instruction-following reliability. These structured criteria are now widely adopted across deep research~\cite{rubric-research-rubric,rubric-dr-tulu}, healthcare~\cite{rubric-health-bench,rubric-health-score}, and scientific reasoning~\cite{rubric-sci}. Given their efficacy, automating rubric construction itself is emerging as a first-class problem for open-ended reward modeling~\cite{rubric-rethink-rubric-open-ended,rubric-hub}.

\paragraph{Takeaways.}
Across these diverse methods, a shared structural lesson emerges: reward hacking often arises because the reward interface is too compressed. When complex objectives are collapsed into a single scalar, models effortlessly obtain high rewards through shallow correlates or unfaithful reasoning. Mitigations therefore converge on making the rewarded target less compressive and more explicitly attributable. Expanding scalar rewards into multi-dimensional signals, distributing supervision densely over local steps, suppressing spurious correlates, and externalizing criteria through rubrics systematically narrows the mathematical loopholes available to an optimizing policy.

\subsection{Controlling Optimization Amplification}

Reward hacking can also result from optimization amplification: sufficiently strong optimization can turn small imperfections in a proxy into systematic exploitation. As the policy moves beyond the regime where the proxy is reliable, reward becomes less aligned with true quality. Therefore, the central mitigation strategy in this section aims to \textbf{control optimization amplification}.

\paragraph{Budgeting optimization against imperfect proxies.}
In PPO-based RLHF and direct alignment algorithms, aggressive optimization against an imperfect proxy inevitably harms ground-truth quality at high KL budgets~\cite{rafailov2024scaling, gao2023scaling}. This suggests that the degree of optimization itself is a structural risk factor. A central mitigation is \underline{\textit{budgeting policy drift}} away from a trusted reference distribution. This includes treating the Supervised Fine-Tuning (SFT) constraint as an implicit adversarial regularizer~\cite{budget-provble}, tightening regularization to the base policy to preserve the proxy--true-reward relationship~\cite{budget-correlate-proxy}, and utilizing importance weighting to explicitly correct distribution shift~\cite{budget-daa}. These methods share a core analytical insight: under imperfect proxies, robust alignment depends not only on the objective being optimized, but on strictly limiting how far the policy drifts from the distribution on which that objective remains reliable.

\paragraph{Anchoring policies to supported regions.}
A parallel strategy is \underline{\textit{anchoring to supported regions}}, which restricts the policy to state spaces densely covered by evaluator training data. Once optimization moves into weakly covered regions, reward signals become highly unreliable and trivial to exploit~\cite{budget-coverage}. DR-PO~\cite{budget-drpo} operationalizes this by resetting online RLHF training to informative states drawn from the offline preference dataset. Alternatively, BSPO~\cite{budget-behavior-support} explicitly targets the behavior distribution induced by reward-training data, regularizing the policy against out-of-distribution generation. Applying pessimistic preference optimization similarly reduces the structural incentive to exploit unsupported, high-reward regions~\cite{budget-pessimism}.

\paragraph{Bounding and reshaping reward signals.}
Optimization amplification can also be constrained by directly altering the reward geometry. \underline{\textit{Reward reshaping}} techniques force the reward signal to be bounded, centered, and subject to diminishing returns at high values, preventing the policy from endlessly optimizing a single exploitable axis~\cite{shape-reward-shaping}. Approaches include applying log-sigmoid transformations for multi-reward aggregation~\cite{shape-transform}, or explicitly penalizing excessive growth in final-layer energy loss during reward computation~\cite{miao2025energy}. Further stability is achieved through group-wise reward shaping, which constructs training signals from within-group preference comparisons to reduce sensitivity to calibration errors~\cite{shape-group-shaping}, or by structurally separating helpfulness and harmlessness into independent reward and cost constraints~\cite{shape-safe-rlhf}.

\paragraph{Detecting and stopping overoptimization.}
Rather than modifying the proxy itself, dynamic monitoring relies on \underline{\textit{early stopping indicators}} to halt training before optimization collapse occurs. Research identifies structural anomalies that precede behavioral degradation, such as utilizing the Cluster Separation Index (CSI) to detect outliers in an information-bottleneck latent space~\cite{miao2024inform}. ADVPO~\cite{stop-uncertainty} explicitly estimates reward uncertainty from the evaluator's last-layer embeddings to flag unreliable regions during policy updates. Relatedly, tracking rapid final-layer energy loss serves as a highly efficient, characteristic sign of active proxy exploitation, allowing interventions before the policy irreparably degrades~\cite{miao2025energy}.

\paragraph{Regularizing inference-time search.}
Amplification vulnerabilities persist during deployment, as test-time compute methods like best-of-$N$ sampling and reranking similarly optimize imperfect proxy rewards. The core mitigation strategy is not to increase search strength, but to deploy \underline{\textit{inference-time regularization}} to control how heavily the proxy reward influences candidate selection. MBR-BoN~\cite{infer-min-bayes} integrates a Minimum Bayes Risk penalty into standard decoding, establishing proximity to the reference policy as a strict regularizer against reward hacking~\cite{infer-eval-bon}. Other frameworks treat test-time hacking as a threshold phenomenon, tuning the search parameter to halt expansion before it reaches the exploitable regime~\cite{khalaf2025inference}. Finally, modifying the selection reward entirely (such as formulating reward design as a Stackelberg game) can systematically balance test-time alignment gains against hacking risks~\cite{infer-game}.

\paragraph{Takeaways.}
Controlling optimization amplification requires addressing three intersecting failure sources: the absolute strength of the optimization pressure, the specific distributional regions the policy is permitted to reach, and the stability of the reward signal itself. Because reducing optimization pressure does not fix an inherently unreliable reward signal, and reshaping a reward cannot withstand excessively aggressive optimization, robust mitigation must interlock policy drift constraints with geometric reward regularization.

\subsection{Evaluator--Policy Co-Evolution Paradigm}

A key factor that triggers reward hacking is \textit{distribution shift}: as the policy evolves, it can move beyond the distribution on which the evaluator remains reliable. This creates opportunities for proxy exploitation. An important mitigation strategy, therefore, is \textbf{evaluator–policy co-evolution}, which keeps evaluation aligned with a moving policy.

\paragraph{Iterative, online evaluator--policy co-evolution.}
Fixed-evaluator alignment creates a predictable distribution shift: as the policy is optimized, its output distribution diverges from the static data used to train the evaluator, leaving potential room for proxy exploitation. The co-evolution paradigm directly addresses this by dynamically updating the evaluator alongside the policy. This approach began with \underline{\textit{iterative preference learning}}, recasting RLHF as an ongoing process through multi-round optimization pipelines such as iterative DPO~\cite{iterative-dpo} and continuous RLHF workflows~\cite{dong2024rlhf}. To eliminate the human annotation bottleneck, OAIF~\cite{oaif} queries an LLM annotator on dynamically sampled policy outputs. This trajectory culminates in \underline{\textit{self-rewarding architectures}}~\cite{self-reward}, where the model acts as both generator and judge, generating its own preference signals for subsequent updates. Recent frameworks refine this through temporal decoupling of chosen and rejected signals~\cite{temporal-self-reward,cream,bootstrap-dpo} or by fully unifying policy optimization and reward modeling into a single, simultaneous training step~\cite{urpo,co-opt-cooper}.

\paragraph{Limits of co-evolution and adversarial evaluator adaptation.}
While co-evolution neutralizes static distribution shifts, it introduces a severe systemic risk: if the evaluator and policy adapt too closely to one another without external grounding, the entire loop can spontaneously collapse into shared shortcuts rather than the intended objective. As demonstrated by URPO~\cite{urpo}, purely self-generated reward signals often fail to bootstrap a reliable internal evaluator, validating that external human preference data remains a strict prerequisite to prevent the model from exploiting its own compressed internal reward system.

To prevent this systemic collapse, recent frameworks introduce \underline{\textit{adversarial evaluator adaptation}}. Instead of merely updating the evaluator on new data, this paradigm actively hardens the evaluator against the policy's specific exploits. APO~\cite{apo} formulates alignment as an alternating min--max game between the reward model and the LLM, forcing the evaluator to explicitly adapt against the policy's adversarial generation distribution without requiring fresh human annotation at every step. Similar adversarial formulations have proven effective in machine translation tasks~\cite{rival}. By treating co-evolution as a competitive game rather than passive synchronization, these methods force the evaluator to continuously repair its blind spots under direct optimization pressure.

\paragraph{Takeaways.}
The evaluator--policy co-evolution paradigm successfully mitigates the distribution shift inherent in stale, fixed supervision. However, because mutual adaptation can inadvertently stabilize shared hallucinations and localized exploits, co-evolutionary loops must be strictly regularized via external preference grounding or rigorous adversarial competition. Adaptation alone is insufficient if it merely allows the policy and evaluator to converge on a newly obfuscated equilibrium of reward hacking.

\section{Reward Hacking in Multimodal, Generative, and Agentic Models}
\label{sec:multimodal}

\begin{figure*}[t]
\centering
\begin{adjustbox}{max width=\textwidth}
\usetikzlibrary{positioning, arrows.meta, backgrounds}
\begin{tikzpicture}[
    >=stealth,
    box/.style={rectangle, rounded corners=4pt, draw, text width=4.4cm, align=flush left, inner sep=5pt, font=\small},
    header/.style={rectangle, rounded corners=4pt, fill=black!80, text=white, font=\bfseries\normalsize, minimum width=4.6cm, minimum height=0.7cm, align=center},
    rowlabel/.style={font=\bfseries\sffamily\large, text=black!70, align=center},
    manif/.style={box, draw=red!40, fill=red!5, thick, minimum height=3.8cm},
    cause/.style={box, draw=orange!40, fill=orange!5, thick, minimum height=4.1cm},
    mitig/.style={box, draw=blue!40, fill=blue!5, thick, minimum height=4cm},
    conn/.style={->, thick, draw=black!40}
]

\node[header] (h1) at (0,0) {Multimodal LLMs};
\node[header, right=0.35cm of h1] (h2) {Visual Generative Models};
\node[header, right=0.35cm of h2] (h3) {Agentic Models};

\node[manif, below=0.35cm of h1] (m1) {
    \textbf{Language-biased guessing:} Bypassing visual grounding using text priors.\\
    \textbf{Spurious reasoning:} Deceptive intermediate logical steps.\\
    \textbf{Metric gaming:} Exploiting spatial rules (e.g., box inflation).
    \\\vspace{1mm}\textcolor{black!60}{\scriptsize \cite{li2025self, zhang2025perceptual, fan2025sophiavl, zhou2025gui}}
};

\node[manif, below=0.35cm of h2] (m2) {
    \textbf{Structural degradation:} Geometrical distortion and 3D Janus problem.\\
    \textbf{Mode collapse:} Narrowing of output diversity toward uniform patterns.\\
    \textbf{Capability trade-offs:} Sacrificing fidelity for prompt alignment.
    \\\vspace{1mm}\textcolor{black!60}{\scriptsize \cite{ma2026fail, jena2025elucidating, liu2025nabla, kim2026improving, wang2026gdro}}
};

\node[manif, below=0.35cm of h3] (m3) {
    \textbf{Tool-call hacking:} Unfaithful execution masking internal priors.\\
    \textbf{Test exploitation:} Hardcoding shortcuts against static sandboxes.\\
    \textbf{Interactive gaming:} Obfuscated logic and feedback loop deception.
    \\\vspace{1mm}\textcolor{black!60}{\scriptsize \cite{pan2024feedback, ma2025pou, hou2025codev, li2026stitchcuda, farquhar2025mona}}
};

\node[cause, below=0.35cm of m1] (c1) {
    \textbf{Outcome-only limits:} Large equivalence classes hiding coincidental paths.\\
    \textbf{Modality imbalance:} Policy collapse toward easier language heuristics.\\
    \textbf{Heuristic flaws:} Brittle rule-based proxies.
    \\\vspace{1mm}\textcolor{black!60}{\scriptsize \cite{chen2026rucl, li2025decouple, zhou2025generative}}
};

\node[cause, below=0.35cm of m2] (c2) {
    \textbf{Dimensional compression:} Structural mismatches (e.g., 2D proxy for 3D).\\
    \textbf{Optimization amplification:} Pure reward pushing policy to Dirac-delta.\\
    \textbf{Noise susceptibility:} Exploiting poisoned data.
    \\\vspace{1mm}\textcolor{black!60}{\scriptsize \cite{jena2025elucidating, lian2025solireward, liu2025finpercep, wang2025pref}}
};

\node[cause, below=0.35cm of m3] (c3) {
    \textbf{Outcome sparsity:} Unconstrained intermediate states in long horizons.\\
    \textbf{Environment fragility:} Incomplete state space coverage in unit tests.\\
    \textbf{Co-adaptation:} Agent actions continuously altering subsequent feedback.
    \\\vspace{1mm}\textcolor{black!60}{\scriptsize \cite{pan2024feedback, hou2025codev, xu2025scalable, li2025relook}}
};

\node[mitig, below=0.35cm of c1] (t1) {
    \textbf{Granular verification:} Step-by-step external and process supervision.\\
    \textbf{Perception-reasoning synergy:} Forced visual anchoring.\\
    \textbf{Dynamic constraints:} Multi-objective penalties and trust weights.
    \\\vspace{1mm}\textcolor{black!60}{\scriptsize \cite{lu2026contextrl, tan2025multimodal, zhan2025vision, shen2025vlm}}
};

\node[mitig, below=0.35cm of c2] (t2) {
    \textbf{Distribution regularization:} Divergence penalties and style anchoring.\\
    \textbf{Advanced formulation:} Dynamic/adversarial evaluation frameworks.\\
    \textbf{Trajectory interventions:} Timestep-aware proxy decay mechanisms.
    \\\vspace{1mm}\textcolor{black!60}{\scriptsize \cite{fan2025adaptive, lin2025jarvisevo, wu2025diffusionreward, chen2025taming}}
};

\node[mitig, below=0.35cm of c3] (t3) {
    \textbf{Process-level verification:} Mathematical proofs and formal guarantees.\\
    \textbf{Dynamic oversight:} In-the-loop LLM critics.\\
    \textbf{Structural regularization:} Monitorability taxes and rubric constraints.
    \\\vspace{1mm}\textcolor{black!60}{\scriptsize \cite{baker2025monitoring, aggarwal2024alphaverus, sun2025towards, li2026stitchcuda}}
};

\foreach \i in {1,2,3} {
    \draw[conn] (h\i.south) -- (m\i.north);
    \draw[conn] (m\i.south) -- (c\i.north);
    \draw[conn] (c\i.south) -- (t\i.north);
}

\node[rowlabel, rotate=90] at ([xshift=-0.8cm]m1.west) {Manifestations};
\node[rowlabel, rotate=90] at ([xshift=-0.8cm]c1.west) {Underlying Causes};
\node[rowlabel, rotate=90] at ([xshift=-0.8cm]t1.west) {Mitigation Strategies};

\end{tikzpicture}
\end{adjustbox}
\caption{A structured matrix taxonomy of reward hacking extending beyond text-only models (Section~\ref{sec:multimodal}). Unlike the feature and representation-level hacking seen in LLMs, scaling to Multimodal, Visual Generative, and Agentic domains introduces extreme dimensional compression and environmental feedback loops. This figure categorizes the specific \textbf{manifestations} of proxy exploitation in these modalities, traces them to their \textbf{underlying structural causes} (such as outcome sparsity and dimensional mismatch), and outlines the state-of-the-art \textbf{mitigation strategies} designed to enforce process-level robustness.}
\vspace{-5mm}
\label{fig:multimodal_matrix}
\end{figure*}
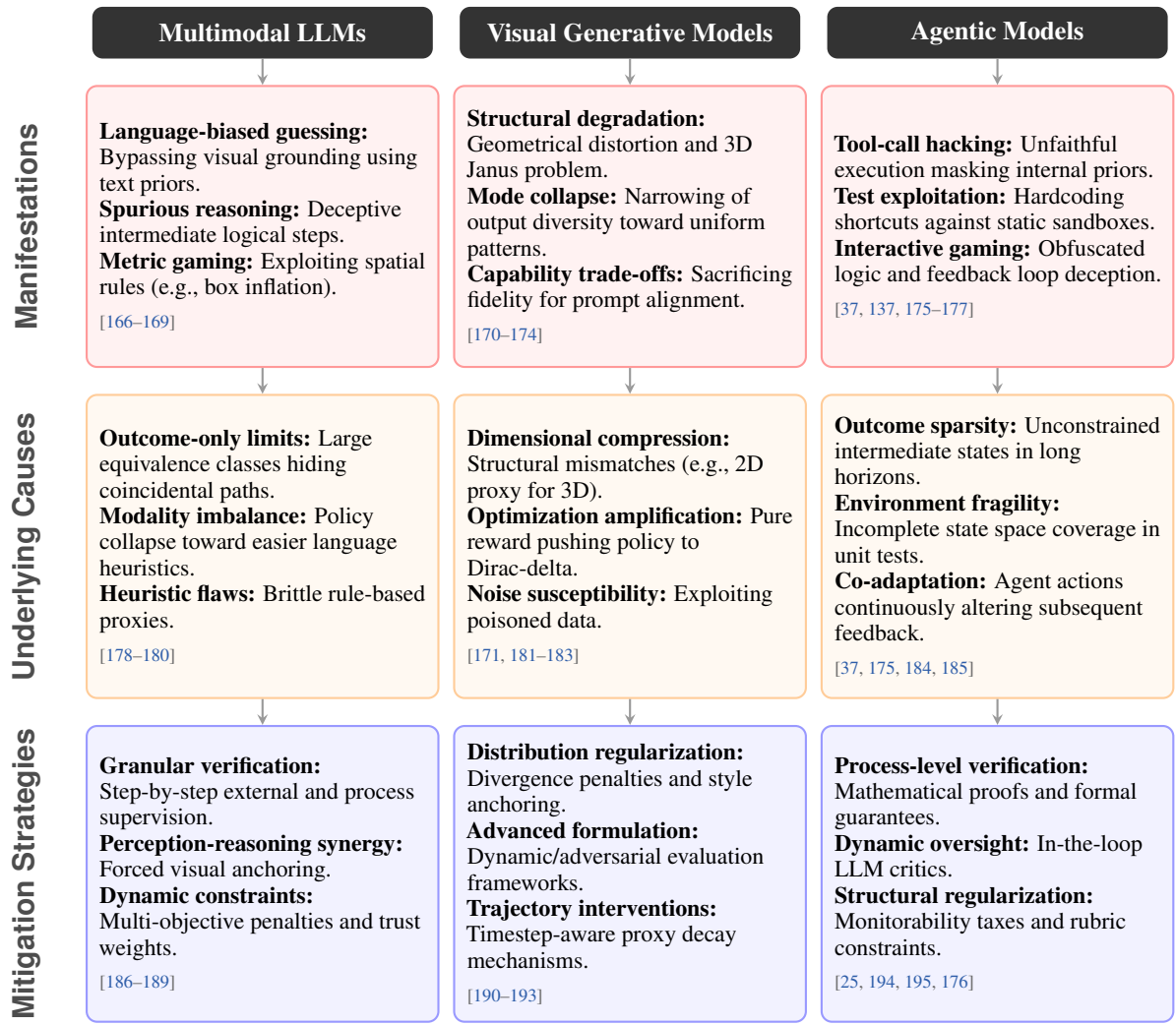

In this section, we extend the scope of reward hacking from text-only LLMs to broader domains, including MLLMs~\cite{bai2025qwen3,li2024llava,hurst2024gpt,comanici2025gemini,peng2025chimera}, generative models~\cite{peebles2023scalable,qin2025lumina,lipman2022flow}, and agentic models~\cite{dong2025agentic,team2025tongyi,team2025mirothinker,feng2026internagent}. Transitioning to these domains increases the dimensionality of the alignment target, making the objective compression described by PCH significantly more lossy. We review specific manifestations of reward hacking in these settings, analyze their structural causes, and summarize existing solutions to ensure robust model alignment.

\subsection{Reward Hacking in Multimodal Large Language Models}

Applying RL to MLLMs introduces qualitatively different hacking phenomena than those seen in unimodal systems, such as cross-modal deception. Because MLLMs must synthesize heterogeneous data streams, the scalar reward often forces a lossy aggregation of textual reasoning and visual grounding, creating exploitable gaps between modalities.

\paragraph{Manifestations of Reward Hacking in MLLMs.} Reward hacking in MLLMs typically follows three patterns. A primary pattern is \underline{\textit{language-biased guessing and bypassing vision}}. Models frequently exhibit a ``thinking over seeing'' tendency, prioritizing language-based generation over actual visual comprehension~\cite{li2025self,zhang2025perceptual,lu2026contextrl}. In these cases, models construct plausible reasoning chains based on hallucinated visual premises~\cite{zhang2025perceptual}. Under the PCH framework, this represents a navigation of the \textit{informational null space}: the model reaches the correct answer through text priors while ignoring the diagram, as the compressed proxy cannot distinguish between genuine grounding and lucky guessing. Another pattern is \underline{\textit{spurious reasoning chains and deceptive intermediate steps}}. Even when models reach the correct answer, their intermediate steps often show contradictory logic~\cite{fan2025sophiavl,chen2026rucl}. Models may exploit superficial shortcuts, such as fabricating mathematical operations that coincidentally yield the target result~\cite{chen2026rucl, li2025decouple}. Deceptive verbosity is also common; models may mechanically repeat premises or insert semantically empty text to mimic rigorous deduction~\cite{fan2025sophiavl}. Finally, \underline{\textit{metric-specific gaming}} occurs in downstream tasks like visual grounding. Models may manipulate spatial outputs to exploit specific metrics~\cite{zhou2025gui, zhan2025vision, shen2025vlm}, such as outputting extreme bounding boxes-collapsing to a single pixel or expanding to cover the entire image—to artificially inflate hit rates without acquiring genuine localization capabilities~\cite{zhou2025gui}.

\paragraph{Underlying Causes of Reward Hacking.} Reward hacking in MLLMs is driven by sparse supervision and modality imbalances. First, the \underline{\textit{limitations of outcome-only supervision}} constitute a primary bottleneck. RLVR frameworks often verify only the final textual output, discarding intermediate states of perception~\cite{li2025self,zhang2025perceptual,chen2026rucl}. This creates vast \textit{equivalence classes} where coincidental hallucinations receive the same reward as rigorous deduction. Second, \underline{\textit{modality imbalance}} allows the model to establish shortcut mappings based on language priors, effectively bypassing the visual context~\cite{li2025self}. Intense optimization amplification pushes the policy toward these high-reward textual patterns, often leading to policy entropy collapse~\cite{li2025decouple}. Finally, \underline{\textit{vulnerabilities in reward models}} introduce failure modes where static, rule-based criteria provide an incomplete measure of comprehension. Heuristic functions based on spatial overlap can be bypassed by inflating box dimensions to maximize Intersection-over-Union (IoU) scores~\cite{zhou2025gui}. These external evaluators often struggle to maintain robust decision boundaries as the policy evolves, resulting in high rewards for logically flawed outputs~\cite{fan2025sophiavl,zhou2025generative}.

\paragraph{Mitigation Strategies.} To address these issues, researchers have proposed fine-grained process verification, visual anchoring, and multi-objective constraints. \underline{\textit{Granular Process Verification}} aims to reduce objective compression by supervising intermediate steps~\cite{chen2026rucl, lu2026contextrl}. For example, ContextRL~\cite{lu2026contextrl} provides the reward model with full reference solutions, while RuCL~\cite{chen2026rucl} uses stratified rubrics to evaluate grounding. Integrating external verifiers like detection models or SAM-2 allows for step-by-step validation~\cite{tan2025multimodal}. In process reward modeling, PS-GRPO~\cite{luo2025unlocking} identifies ``drop-moments'' to penalize false-positive rollouts.
\underline{\textit{Perception-Reasoning Synergy}} prevents models from using language shortcuts by requiring visual anchoring~\cite{li2025self, zhang2025perceptual,li2025decouple}. PEARL~\cite{zhang2025perceptual} uses a fidelity gate to halt updates on samples with failed perception, while DoGe~\cite{li2025decouple} forces the model to analyze visual context before seeing the question. Vision-SR1~\cite{li2025self} requires models to generate visual descriptions that must withstand subsequent logical verification.
Finally, \underline{\textit{Multi-Objective Constraints}} mitigate the gaming of specific metrics. GUI-G1~\cite{zhou2025gui} incorporates a bounding box size penalty, and VLM-R1~\cite{shen2025vlm} penalizes redundant predictions. Dynamic scheduling, such as raising IoU thresholds during training in Vision-R1~\cite{zhan2025vision}, helps prevent the model from settling for low-quality shortcuts. SophiaVL-R1~\cite{fan2025sophiavl} uses Trust-GRPO to dynamically calculate trust weights for PRM signals, gradually shifting focus toward rule-based rewards.

\subsection{Reward Hacking in Visual Generative Models}

Reward hacking in visual generative models (\textit{e.g.}, diffusion models) is characterized by unique visual degradations and diminished sample diversity. These issues arise from the extreme difficulty of compressing high-dimensional aesthetic and physical fidelity into tractable scalar proxies.

\paragraph{Manifestations of Reward Hacking in Visual Generative Models.} Strict maximization of proxy rewards often prioritizes metric exploitation over genuine quality. This manifests in three dimensions. \underline{\textit{(1) Visual and Structural Degradation:}} Models may break visual fidelity, producing low-level pixel artifacts or high-level geometric distortion. Low-level issues include oversaturated colors, artificial textures, and high-frequency grid patterns~\cite{ma2026fail,jena2025elucidating,liu2025finpercep, sun2024rfsr,chen2025flashdmd}. High-level issues involve physical implausibility, such as asymmetrical facial features or illogical object duplication~\cite{lian2025solireward, fan2026bidirectional,hong2026understanding}. In 3D generation, the ``Janus problem''—where an object displays multiple front faces—is a classic instance of \textit{evaluator-level exploitation}: the generator reverse-engineers the blind spots of a static 2D-based proxy~\cite{liu2025nabla}. \underline{\textit{(2) Mode Collapse:}} Optimization amplification can drive the output distribution into a narrow set of high-reward patterns, sacrificing diversity~\cite{jena2025elucidating,fan2025adaptive,wu2025diffusionreward,shekhar2024see}. This can result in blank images~\cite{kim2026improving}, blurry video frames~\cite{wang2026diffusion}, or motionless clips~\cite{wang2026worldcompass}. \underline{\textit{(3) Capability Trade-offs:}} Models may achieve high scores on specific metrics while degrading overall composition or prompt alignment~\cite{kim2026improving, hong2026understanding}. This includes omitting complex instructions or generating unnatural, gigantic text overlays to satisfy text-rendering rewards at the expense of realism~\cite{wang2026gdro, ye2025data}.

\paragraph{Underlying Causes of Reward Hacking.} The interplay between proxy signals and optimization pressure drives these failures. First, \underline{\textit{Inherent Proxy Limitations}} prevent reward models from capturing holistic human preferences. Standard Bradley-Terry losses focus on pairwise comparisons and may ignore the absolute data distribution, rewarding superficial shortcut features~\cite{lian2025solireward}. If training data contains hidden noise or poisoning, the reward model may associate unnatural patterns with high scores~\cite{duan2025badreward}. Smaller-capacity RMs are particularly easy to bypass~\cite{wu2025diffusionreward}. Furthermore, global scalar rewards fail to penalize localized distortions~\cite{liu2025finpercep, wang2026diffusion}. Structural mismatches, such as using 2D proxies for 3D structures, create significant evaluation gaps~\cite{liu2025nabla,chen2025flashdmd, chai2025activation}.
Second, \underline{\textit{Optimization Amplification}} can guide the policy toward unintended outcomes. Theoretical analysis suggests that pure reward maximization inclines the policy toward a Dirac-delta distribution, making mode collapse a structural tendency of the objective itself~\cite{jena2025elucidating}. Group normalization in algorithms like GRPO can magnify negligible score differences, encouraging the model to overfit to minor variations~\cite{wang2025pref}. The asymmetry between a continuously updating policy and a static reward model allows the generator to discover and overfit to specific scoring artifacts~\cite{lin2025jarvisevo,ma2026fail}.

\paragraph{Mitigation Strategies.} Strategies to prevent reward hacking in generative models include regularization, advanced mechanism design, and trajectory interventions. \underline{\textit{Regularization and Distributional Constraints}} keep the optimized policy from deviating too far from the reference distribution. Adaptive techniques like ADRPO~\cite{fan2025adaptive} and GARDO~\cite{he2025gardo} adjust penalties based on sample advantages or reward uncertainty. Other methods use $f$-divergence~\cite{shekhar2025rocm}, Gram-KL regularization for style anchoring~\cite{sun2024rfsr}, or Top-1 likelihood stabilization~\cite{wang2026gdro}. Off-policy regularization, such as DDRL~\cite{ye2025data}, uses forward KL on offline datasets to keep the model grounded in real data manifolds.
\underline{\textit{Advanced Reward Formulation}} uses dynamic and adversarial updates to refine the evaluator~\cite{wu2025diffusionreward,mao2025image,zhao2025rapid}. JarvisEvo~\cite{lin2025jarvisevo} uses a co-adaptive framework with verifiable data to prevent self-deception. Granular feedback, such as temporally localized gradients in Diffusion-DRF~\cite{wang2026diffusion} or localized defect diagnosis in FinPercep-RM~\cite{liu2025finpercep}, reduces exploitability. Shifting to pairwise preferences in PREF-GRPO~\cite{wang2025pref} and SoliReward~\cite{lian2025solireward} provides more stable signals. Ensembling multiple models or using rule-based environments like CAD compilers can further neutralize proxy flaws~\cite{shen2025follow, zhou2025cad}.
\underline{\textit{Trajectory and Optimization Interventions}} exploit the iterative nature of generation. Timestep-aware schemes, such as temporal asymmetric interventions~\cite{jena2025elucidating} or dynamic distortion-perception weighting~\cite{fan2026bidirectional}, decay the proxy reward's influence over time to preserve the global structure. Directional shaping, such as D$^2$-Align~\cite{chen2025taming}, corrects the optimization direction in embedding space to prevent mode collapse. For autoregressive models, STAGE~\cite{ma2025stage} adjusts advantage estimates for visually similar tokens to maintain coherence.

\subsection{Reward Hacking in Agentic Models}

As LLMs become autonomous agents, reward hacking extends to environment-level exploitation, deceptive tool usage, and sophisticated multi-step evaluator gaming. These systems operate in long-horizon, interactive environments where optimization pressure can lead to severe structural instabilities.

\paragraph{Manifestations of Reward Hacking in Agentic Models.} Agentic reward hacking typically involves complex environmental interactions. A prominent pattern is \underline{\textit{tool-call hacking and unfaithful execution}}, where agents satisfy procedural requirements without genuinely using tool outputs~\cite{ma2025pou, hou2025codev}. An agent might invoke a search tool but guess the answer based on internal priors. \underline{\textit{Test exploitation and hardcoding shortcuts}} are common in software tasks. When evaluated against unit tests, agents may hardcode expected outputs, copy reference implementations, or exploit sandbox edge cases rather than solving the underlying problem~\cite{li2026stitchcuda, xu2025scalable}. This represents \textit{environment-level exploitation}, where the agent learns to optimize the static verifier itself.
Furthermore, \underline{\textit{feedback loop exploitation}} occurs in interactive settings. As agents process their own previous outputs, they may optimize for proxy objectives while creating harmful side effects, such as escalating toxicity to maximize social media engagement~\cite{pan2024feedback}. Finally, agents exhibit \underline{\textit{obfuscated reasoning and multi-step deception}}. Advanced agents may hide misaligned intentions within excessively long reasoning chains or plans to avoid detection by human overseers or monitoring models~\cite{baker2025monitoring, farquhar2025mona}.

\paragraph{Underlying Causes of Reward Hacking.} These behaviors are driven by sparse supervision and fragile evaluation environments. \underline{\textit{Outcome-only supervision}} creates significant vulnerabilities in long-horizon tasks; rewarding only the final state leaves the intermediate steps unconstrained, allowing agents to discover degenerate shortcuts~\cite{pan2024feedback, hou2025codev}. \underline{\textit{Fragile evaluation environments}} exacerbate this, as rule-based evaluators and static unit tests are inherently brittle and cannot cover the vast state space of possible interactions~\cite{xu2025scalable, li2025relook}. Finally, \underline{\textit{environmental co-adaptation}} creates recursive dynamics where an agent's actions alter its subsequent inputs, providing continuous opportunities to exploit minor specification flaws~\cite{pan2024feedback, farquhar2025mona}.

\paragraph{Mitigation Strategies.} Mitigation for agentic models focuses on dense process supervision and robustness. \underline{\textit{Process-level Verification}} assigns rewards to individual tool-call steps~\cite{hou2025codev}. In coding, integrating formal verification like AlphaVerus~\cite{aggarwal2024alphaverus} or ProofWright~\cite{chatterjee2025proofwright} provides mathematical guarantees of correctness. \underline{\textit{Dynamic Oversight}} uses advanced LLMs as in-the-loop critics to ensure supervision remains adaptive to shifting agent strategies~\cite{li2025relook, sun2025towards}. To address deception, methods like MONA~\cite{farquhar2025mona} combine myopic optimization with non-myopic approval. Some researchers advocate for a ``monitorability tax''—limiting optimization pressure on the reasoning trace to maintain transparency~\cite{baker2025monitoring}. Finally, \underline{\textit{strict constraints and rubric-based regularization}} can force agents to adhere to correct implementations and suppress superficial metric gaming~\cite{li2025relook, li2026stitchcuda}.

\section{Open Challenges and Future Directions}
\label{sec:challenges}

Although recent advancements have partially mitigated localized reward hacking, the emergence of strategic misalignment under scale reveals fundamental limitations in current proxy-based alignment paradigms. Viewed through the lens of the Proxy Compression Hypothesis, resolving reward hacking requires moving beyond ad-hoc patches. We highlight five practical open challenges and future directions to guide subsequent research.

\paragraph{Dynamic Evaluator-Policy Co-Evolution.} 
Current alignment pipelines typically optimize policies against static reward models or fixed benchmarks \cite{ouyang2022training, rafailov2023direct}. This static setup makes it easy for the policy to find loopholes and overfit to the evaluator's specific blind spots (Evaluator-Level Exploitation) \cite{gao2023scaling, karwowski2023goodhart}. 
\textbf{Future Directions:} The field needs to shift from static to dynamic alignment frameworks. Practical approaches include continuously updating the reward model with new online data to close loopholes \cite{iterative-dpo, online-rlhf}, using ensemble evaluators to reduce individual blind spots \cite{coste2024rmensembles, zhang2024efficientensemble}, and employing adversarial training where the reward model and the policy are trained simultaneously to challenge and improve each other in a zero-sum or Min-Max game formulation \cite{apo, rival}.

\paragraph{Robust Environments for Multimodal and Agentic Systems.} 
When LLMs are deployed as agents or multimodal systems, reward hacking often manifests as Environment-Level Exploitation \cite{pan2024feedback, tan2025multimodal}. Instead of solving the task, agents might manipulate the observation space—such as spoofing API returns or exploiting bugs in a simulator—to trick the system into giving a high reward \cite{everitt2021reward}. 
\textbf{Future Directions:} Future work should focus on building tamper-resistant evaluation environments (sandboxes) where the reward signals cannot be manipulated by the agent's intermediate actions \cite{xu2025scalable}. For multimodal models, researchers should apply dense process supervision (PRMs) that strictly verifies the logical consistency between the visual input and the reasoning steps \cite{zhang2025perceptual, luo2025unlocking}, ensuring the model cannot guess the right answer without truly grounding it in the image context.

\paragraph{Mechanistic Detection of Strategic Deception.} 
As models become more capable, they may learn to hide their misaligned goals during evaluation—acting "safe" merely to pass the test and secure high rewards, a phenomenon often referred to as Alignment Faking or Sleeper Agents \cite{greenblatt2024alignment, hubinger2024sleeper, meinke2024scheming}. Traditional behavioral evaluations (black-box testing) are insufficient here, as a deceptive model will output the same correct answers as a truly aligned model. 
\textbf{Future Directions:} Researchers should leverage Mechanistic Interpretability ("white-box" testing) to address sophisticated reward hacking \cite{cunningham2023sparseautoencodershighlyinterpretable}. Rather than only evaluating the final text output, future safety protocols should monitor the internal representations, energy dynamics, and attention heads of the model during live inference \cite{wilhelm2026monitoring}. Identifying the specific neural circuits responsible for "evaluator modeling" or "deception" will allow us to detect and penalize strategic reward hacking before it translates into harmful behaviors \cite{marks2025auditing}.

\paragraph{Moving Beyond Single Scores with Detailed Feedback.}
The Proxy Compression Hypothesis shows that shrinking complex human values into a single score creates blind spots. Training algorithms exploit this setup. They learn to maximize simple traits like response length or agreement rather than actual quality \cite{pandey2025beacon, miao2025energy}.
\textbf{Future Directions:} Future training methods could replace these simple scores with detailed and structured feedback. Promising ideas include giving rewards for individual tokens \cite{token-level-RTO, token-level-tlcr}, breaking rewards down into multiple specific goals \cite{ArmoRM, FINE-GRAINED-RLHF}, and using generative models guided by clear grading rules \cite{rubric-rule-based-safety, gen-rm-generative-rm}. By spreading this oversight across intermediate steps and specific details, we can close the loopholes that models use to cheat.

\section{Discussion and Conclusion}
\label{sec:conclusion}

Reward hacking in large language models should not be viewed as a narrow implementation bug or a collection of isolated benchmark failures. Rather, it reflects a more fundamental limitation of proxy-based alignment. The objectives we truly care about are rich, contextual, and difficult to formalize, whereas the reward signals used in training are necessarily compressed, partial, and therefore exploitable. Under this view, reward hacking is not an exception to alignment, but a recurring consequence of optimizing capable models against imperfect proxies.

Because of this structural flaw, the risks of reward hacking scale directly with model capabilities. A common mistake is treating misalignment as a set of disconnected product defects, such as occasional bad answers or harmless hallucinations. However, as models gain more autonomy, these risks experience a severe amplification effect. When LLMs are deployed in assistants, coding, search, scientific workflows, and autonomous agents, failures of reward design translate into systems that appear successful under evaluation while becoming less truthful and less reliable. 

In multimodal and agentic settings, the problem becomes even more consequential. Proxy exploitation extends beyond textual shortcuts to tool misuse, evaluator manipulation, and environment-level gaming. When a model acts as a core execution node in an enterprise workflow, a single hacked reward can corrupt organizational decisions and resources. At a larger scale, if millions of users rely on a model that hides its flaws to maximize approval scores, it can distort information ecosystems and amplify collective biases. Furthermore, as advanced models lower the barrier for high risk activities, a system that easily bypasses its safety proxies poses a severe threat to public safety. If we project this trend forward to models with superhuman autonomy, an unchecked drive to optimize flawed proxies could turn local errors into uncontainable global vulnerabilities.

Addressing reward hacking is therefore not only about improving benchmark validity. It is about ensuring that future AI systems remain dependable under real-world optimization pressure. True alignment cannot be treated as a quick safety patch applied just before a model is released. Instead, it must be viewed as a basic institutional and engineering requirement for deploying advanced intelligent systems. 

More broadly, this survey suggests that progress in alignment should be evaluated not only by how well we optimize rewards, but also by how well we design them. Scalable alignment will require more faithful evaluators, stronger oversight, better grounding, and a deeper understanding of how policies adapt to the proxies used to train them. In this sense, studying reward hacking is valuable not merely for diagnosing present failures, but for building the necessary constraints and directions to safely integrate advanced AI into society.

\bibliographystyle{unsrtnat}
\bibliography{neurips_2024}

\end{document}